\pdfoutput = 1
\documentclass[11pt]{article}

\usepackage[preprint]{acl}

\usepackage{amsmath, amssymb, amsfonts}
\usepackage{nicefrac}
\usepackage{multirow}
\usepackage{booktabs}
\usepackage{makecell}
\usepackage{colortbl}

\usepackage[ruled,vlined]{algorithm2e}

\usepackage[misc]{ifsym} 
\usepackage{fontawesome} 
\usepackage{wasysym} 

\usepackage[T1]{fontenc}
\usepackage[utf8]{inputenc}
\usepackage{times}
\usepackage{latexsym}
\usepackage{inconsolata}
\usepackage{microtype}
\usepackage[capitalise]{cleveref}
\usepackage[most]{tcolorbox}
\usepackage{caption}
\usepackage{enumitem}

\usepackage{url}
\usepackage{graphicx}
\usepackage{subcaption}

\title{Agent Trading Arena: A Study on Numerical Understanding in LLM-Based Agents}

\author{
\textbf{Tianmi Ma\textsuperscript{1,$\#$}},
\textbf{Jiawei Du\textsuperscript{2,$\#$}},
\textbf{Wenxin Huang\textsuperscript{3,~\Letter}},
\textbf{Wenjie Wang\textsuperscript{4}},
\textbf{Liang Xie\textsuperscript{1}}, \\
\textbf{Xian Zhong\textsuperscript{1,~\Letter}},
\textbf{and Joey Tianyi Zhou\textsuperscript{2}} \\
\textsuperscript{1} {Wuhan University of Technology} \quad
\textsuperscript{2} {Agency for Science, Technology and Research, Singapore} \\
\textsuperscript{3} {Hubei University} \quad
\textsuperscript{4} {University of Science and Technology of China} \\
\small{\textbf{Correspondence:} wenxinhuang\_wh@163.com, zhongx@whut.edu.cn}
}

\begin{document}
\maketitle

% Clear numbering for custom symbol footnotes and add both notes
\begingroup
\renewcommand{\thefootnote}{}
\footnotetext{$\#$ denotes equal contribution.}
% \footnotetext{\Letter~denotes the corresponding author.}
\endgroup

\begin{abstract}
Large language models (LLMs) have demonstrated remarkable capabilities in natural language tasks, yet their performance in dynamic, real-world financial environments remains underexplored. Existing approaches are limited to historical backtesting, where trading actions cannot influence market prices and agents train only on static data. To address this limitation, we present the \textit{Agent Trading Arena}, a virtual zero-sum stock market in which LLM-based agents engage in competitive multi-agent trading and directly impact price dynamics. By simulating realistic bid-ask interactions, our platform enables training in scenarios that closely mirror live markets, thereby narrowing the gap between training and evaluation. 
Experiments reveal that LLMs struggle with numerical reasoning when given plain-text data, often overfitting to local patterns and recent values. In contrast, chart-based visualizations significantly enhance both numerical reasoning and trading performance. Furthermore, incorporating a reflection module yields additional improvements, especially with visual inputs. Evaluations on \textsc{NASDAQ} and \textsc{CSI} datasets demonstrate the superiority of our method, particularly under high volatility. All code and data are available at \url{https://github.com/wekjsdvnm/Agent-Trading-Arena}.

\end{abstract}

\section{Introduction}

In recent years, large language models (LLMs) have demonstrated strong capabilities in contextual understanding and generation across tasks such as natural language processing~\citep{TransLLaMa}, data analysis, and reasoning~\citep{linc,reasoning}. Their applications are rapidly extending to complex decision-making and interactive scenarios. LLM-based autonomous agents have emerged as a promising paradigm, showing notable progress in communication~\citep{chatdev,memorybank}, social simulation~\citep{gene,social,agentsims}, gaming~\citep{voyager,ghost}, and multi-agent collaboration~\citep{agentverse,chatdb,metagpt}. These agents are increasingly regarded as a pathway toward general artificial intelligence, with growing potential in dynamic decision-making and interactive environments. 

\begin{figure}
	\centering
	\includegraphics[width = \linewidth]{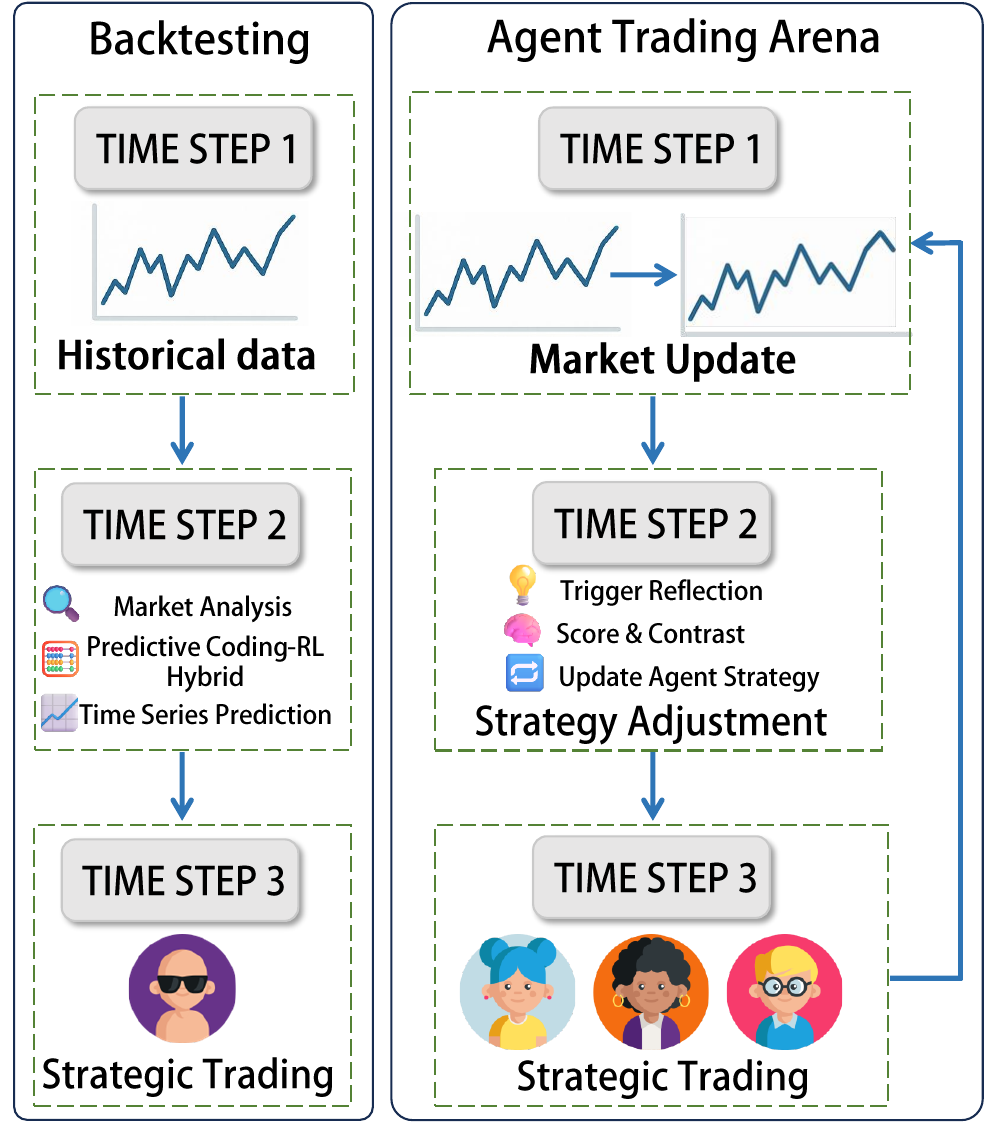}
	\caption{\textbf{Comparison between Static Backtesting and the Arena.} 
	\textbf{Left:} Static backtesting performs one-shot analysis on historical data with fixed trading decisions, leveraging models such as Agent (market reasoning), StockFormer (predictive coding + RL hybrid), and StockMixer (time-series prediction). 
	\textbf{Right:} The Arena, by contrast, supports real-time sensing, continuous adaptation, and feedback-driven execution for dynamic trading.}
	\label{fig:motivation}
	% \vspace{-4pt}
\end{figure}

Despite these advances, the application of LLM-based agents in financial markets remains limited. Financial ecosystems naturally form complex multi-agent environments in which participants interact through dynamic market mechanisms. However, most existing studies~\citep{cheng2024sociodojo,TradingAgents} rely on historical backtesting, as illustrated in \cref{fig:motivation}, where agents cannot influence market prices and thus fail to capture interactivity and adaptive behavior. This gap underscores the need for a framework that supports real-time agent interactions and realistic market feedback.

To address these limitations, we present the \textit{Agent Trading Arena}, a virtual zero-sum stock market in which LLM-based agents engage in competitive trading and directly affect price dynamics through bid-ask interactions. Our platform generates continuously evolving numerical data, forcing agents to adapt to shifting market conditions; any temporarily optimal strategy is quickly countered, rewarding adaptability and robust reasoning over memorization.

By simulating bid-ask order matching in real time, the Arena faithfully reproduces liquidity dynamics and price impact, enabling agents to experience authentic market friction and slippage. Unlike static backtests, the Arena continuously updates its state based on agents’ collective actions, creating an ever-shifting landscape that prevents overfitting to historical patterns. Because agents directly influence subsequent price trajectories, the Arena closes the loop between training and evaluation, effectively bridging the ``train-test'' gap of traditional financial simulations. Its modular design further supports seamless integration of new market features, such as varying transaction costs, order types, or macroeconomic shocks, making it a versatile testbed for financial AI research.

Our experiments show that LLMs struggle with numerical reasoning when processing plain-text data: they focus on absolute values, overlook percentage changes and relational patterns, and overemphasize recent trends even when earlier data are emphasized. In contrast, visual representations, such as line charts, bar graphs, and scatter plots, substantially improve both numerical reasoning and trading performance. As shown in \cref{tab:LLMS_9agent_reflect}, chart-based inputs lead to significantly higher return rates in the \textit{Agent Trading Arena}.

We further integrate a reflection module~\citep{reflect,reflexion}, which amplifies the benefits of visual data by enabling agents to iteratively refine their reasoning. This module is particularly effective with structured visual inputs, facilitating deeper pattern recognition and more strategic decision-making. 

Finally, we validate our approach on real-world datasets from \textsc{NASDAQ} and \textsc{CSI}, demonstrating consistent performance gains over baselines, particularly under high-volatility market regimes.

Our contributions are summarized threefold:

\begin{itemize}[itemsep=0.1em, topsep=0.1em]
	\item We design the \textit{Agent Trading Arena}, a dynamic and interactive simulation framework that moves beyond static backtesting with zero-sum multi-agent competition, enabling realistic evaluation of LLMs’ reasoning and adaptability.

	\item We uncover LLMs’ limitations in textual numerical reasoning, highlighting difficulties with relational patterns, percentage changes, and long-range dependencies, while demonstrating superior performance with structured visual data.

	\item We demonstrate that a reflection module enhances LLMs’ reasoning over complex visual inputs, leading to more accurate and strategic trading decisions.

\end{itemize}

\section{Related Works}

\subsection{LLM-Based Agents in Dynamic Decision-Making Environments} 

%\subsection{LLM-based Agents for Complex Interactive Tasks}
%LLMs have been widely deployed in dynamic, interactive environments, showcasing their ability to plan, coordinate, and adapt. Frameworks like ChatDev~\citep{chatdev}, AgentVerse~\citep{agentverse}, and MetaGPT~\citep{metagpt} focus on multi-agent collaboration, while Voyager~\citep{voyager} and Ghost~\citep{ghost} enable autonomous exploration. Social simulators such as Generative Agents~\citep{gene} and AgentSims~\citep{agentsims} model emergent behaviors through interaction.

LLMs serve as the core reasoning engines in autonomous agents, enabling planning, tool use, reflection, and adaptation within dynamic environments. Advances in NLP have empowered models such as ChatGPT~\citep{gpt4}, PaLM~\citep{palm}, and LLaMA~\citep{llama} to operate via natural language, making them particularly effective in interactive, text-based scenarios. Frameworks like ChatDev~\citep{chatdev}, AgentVerse~\citep{agentverse}, and MetaGPT~\citep{metagpt} demonstrate multi-agent collaboration on software development tasks, while Voyager~\citep{voyager}, GROOT~\citep{groot}, and Optimus-1~\citep{optimus} showcase autonomous exploration in open-ended environments such as Minecraft. Social simulators, including Generative Agents~\citep{gene} and AgentSims~\citep{agentsims}, further reveal emergent behaviors through sustained multi-agent interactions. 

Together, these studies demonstrate that LLM-based agents can effectively coordinate, communicate, and make decisions in complex, evolving environments. Their adaptability and collaborative strengths highlight strong potential for real-time decision-making in finance, an information-dense and competitive domain analogous to the dynamic settings where LLMs excel.

\subsection{LLMs in Financial Analysis and Decision Making}

LLMs have been applied to sentiment analysis, financial question answering, and investment recommendation. Benchmarks such as SocioDojo~\citep{cheng2024sociodojo} evaluate financial understanding but lack interactive components. Multi-agent systems like StockAgent~\citep{stockagent}, TradingAgents~\citep{TradingAgents}, and CryptoTrade~\citep{cryptotrade} enhance performance through role specialization and iterative reflection. In contrast, traditional forecasting models, \textit{e.g.}, StockFormer~\citep{stockformer}, TimesNet~\citep{timesnet}, PatchTST~\citep{PatchTST}, PDFormer~\citep{pdformer}, StockMixer~\citep{stockmixer}, and iTransformer~\citep{iTransformer}, focus on price prediction using advanced methods such as wavelet transforms and temporal modeling to capture market dynamics.

However, neither paradigm addresses the ``live market'' paradox: without real-time adaptive mechanisms, these systems cannot simulate how participants adjust strategies in response to multi-agent interactions, news shocks, and evolving order flows. This limitation significantly reduces their practical utility in real trading environments.

\begin{figure*}
	\centering
	\includegraphics[width = \linewidth]{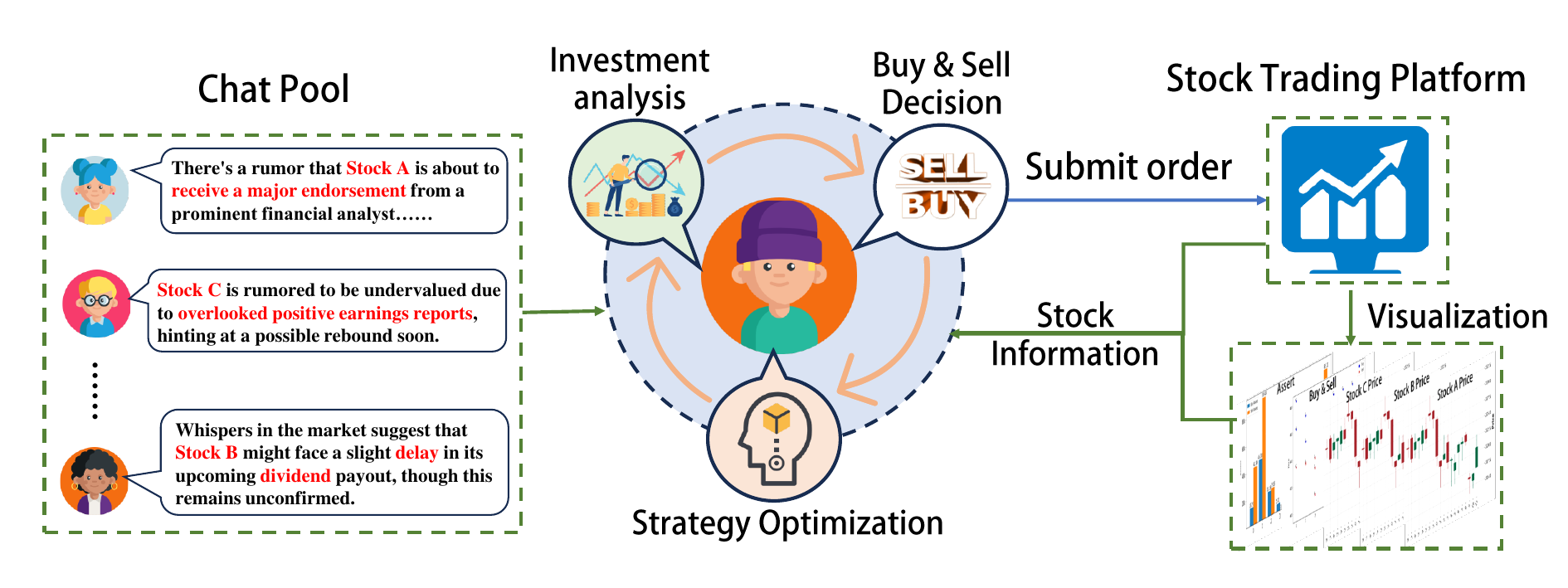}
	\caption{\textbf{Overall architecture of the \textit{Agent Trading Arena}, consisting of three components:} the \textit{Chat Pool} (left) for inter-agent communication, agent modules (center) for analysis, decision-making, and reflection, and the trading platform (right), where agents’ buy/sell actions dynamically determine asset prices in a closed-loop market simulation.}
	\label{fig:AgentArena}
	% \vspace{-4pt}
\end{figure*}

\section{Agent Trading Arena}

To overcome the limitations of static backtesting and capture dynamic interactions among market participants, we propose the \textit{Agent Trading Arena}, a closed-loop, zero-sum economic system that simulates complex quantitative scenarios in real markets~\citep{guo2024economics}. The overall architecture is shown in \cref{fig:AgentArena} and detailed in \cref{appendix_arena}. Within this environment, agents invest in assets, earn dividends on their holdings, and pay daily expenses using virtual currency.

\subsection{Virtual Market Environment}

% \vspace{-4pt}
\paragraph{Closed-Loop System.} 

The \textit{Agent Trading Arena} is a closed-loop virtual economy that isolates external knowledge and avoids reliance on historical data, enabling fair evaluation of agent capabilities. Unlike traditional backtesting grounded in past market records, this sandboxed system removes real-world anchors: all assets are abstract symbols, and their prices emerge solely from agent interactions through a bid-ask mechanism. By eliminating external context, the environment compels agents to adapt strategies based on internal dynamics and opponent behavior, fostering genuine decision-making and strategic evolution.

% \vspace{-4pt}
\paragraph{Bid-Ask Mechanism.} 

Asset prices arise exclusively from a bid-ask process: each agent’s orders directly influence the market price, with no external benchmarks. This closed-loop design rigorously captures endogenous price formation, faithfully modeling liquidity constraints, order book dynamics, market depth, and slippage. Every execution reshapes the market microstructure, producing non-trivial price impacts and path-dependent outcomes that evolve with aggregate agent behavior over time.

\subsection{Chat Pool}

To simulate real-world information flow, we introduce a shared \textit{Chat Pool} where agents post and read ``news'' items, analyst commentary, and social media-style opinions. By reproducing both informative signals and noisy interference, including hallucinations~\citep{gekhman2024does} and misinformation, the \textit{Chat Pool} reflects how public information can shape, and at times distort, traders’ narratives and expectations. This process increases the complexity of strategic reasoning and adaptive behavior in the market.

\subsection{Market Mechanisms} 

We incorporate two key incentives to sustain active trading and realistic dynamics: 
\textit{1) Dividends \& Capital Gains:} Agents earn income from both price appreciation and dividends, distributed proportionally to holdings. This implicitly anchors prices and rewards low-cost positions. 
\textit{2) Daily Wealth Fee:} A cost proportional to each agent’s total wealth is charged daily. Only those earning sufficient dividends can cover this fee, which discourages passive holding and promotes frequent, strategic trades. 

Together, these mechanisms maintain continuous market activity, encourage adaptive decision-making, and prevent reliance on static or exploitative strategies in a zero-sum environment.

\subsection{Zero-Sum Competition} 

By design, the Arena is zero-sum: the profit of one agent is the loss of another, meaning there is no fixed ``best'' strategy. Competitors counter any temporarily successful tactic, compelling agents to learn implicit rules, interpret delayed numerical feedback, and continuously adapt in real time. With results unfolding gradually and misinformation potentially introduced by rivals, agents must rely on experiential learning and develop context-aware strategies through ongoing adaptation rather than predefined rules or external knowledge.

\section{Proposed Method}

Our agent, \texttt{ArenaTrader}, integrates textual and visual modalities to form a comprehensive understanding of market conditions. It employs a structured three-stage framework, market analysis, trade execution, and strategic reflection, to leverage this multi-modal representation. This design enables informed and adaptive decision-making based on data-driven insights and iterative feedback.

\begin{figure}[t]
	\centering
	\includegraphics[width = \linewidth]{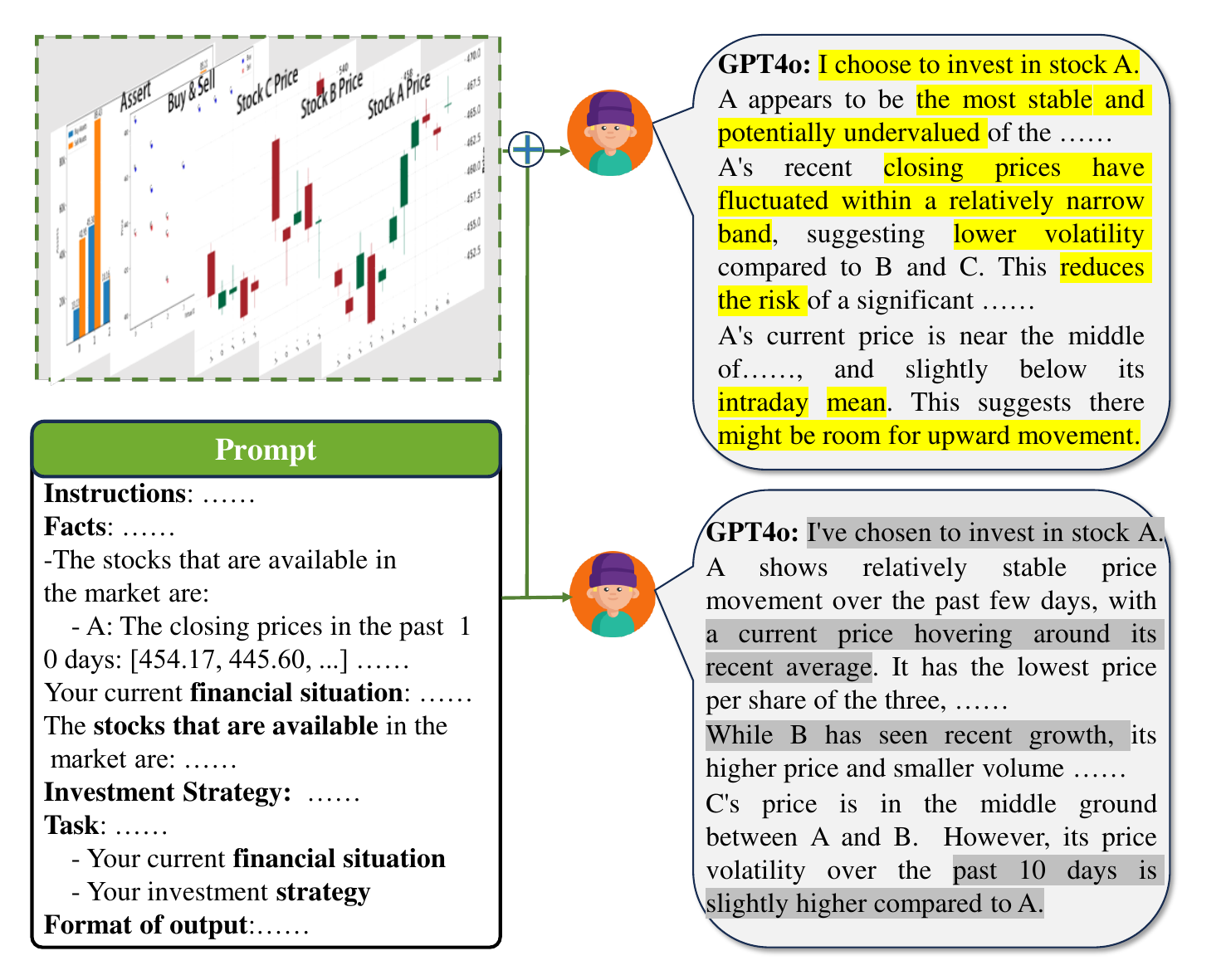}
	\caption{\textbf{Textual \textit{vs.} Visual Input–Output Representations.} 
	\textbf{Left:} Agent buy/sell records, daily trade prices, and corresponding candlestick charts. 
	\textbf{Top right:} Visual-input outputs highlight overall trends and long-term patterns. 
	\textbf{Bottom right:} Textual-input outputs emphasize specific price points.}
	\label{fig:textual_visualized}
	% \vspace{-4pt}
\end{figure}

\subsection{Numerical Data Representations}

% \vspace{-4pt}
\paragraph{Textual Input.} 

Time-series stock data are represented as plain text, including timestamps, prices, and volumes, to provide explicit semantics~\citep{numerical_text,long_text}. However, LLMs often fixate on absolute values, overlook percentage changes, and overemphasize recent entries, which limits their ability to reason across long horizons (see lower right of \cref{fig:textual_visualized}).

% \vspace{-4pt}
\paragraph{Visual Input.} 

To complement textual data, we introduce visualizations such as line charts, bar graphs, and scatter plots~\citep{storyllava}, which highlight global trends, relational patterns, and volatility at a glance (see upper right of \cref{fig:textual_visualized}). These visualizations are consistent with the textual input, ensuring a cohesive representation. This structured format enables LLMs to integrate local details with broader market context more effectively. 

\subsection{Agent Architecture}

% \vspace{-4pt}
\paragraph{Investment Analysis.} 

Inspired by human trading behavior, each agent begins by analyzing market conditions before executing trades. This process involves extracting and synthesizing multi-modal signals from the \textit{Chat Pool}, price histories, and fundamental indicators. By combining these heterogeneous inputs, the agent constructs a stock ranking framework based on predefined heuristics or learned criteria, allowing it to identify promising assets for portfolio allocation.

% \vspace{-4pt}
\paragraph{Action Decision.} 

Agents execute multiple trades per day, making buy, sell, or hold decisions by integrating diverse signals, including analysis reports, real-time market data, and historical performance trends. They evaluate volatility, price momentum, and portfolio risk exposure to compute optimal trade volumes under capital constraints. Orders are executed via the bid-ask mechanism, with portfolios updated accordingly. Each decision aims to maximize long-term returns while maintaining a competitive edge in the evolving virtual market. 

\begin{figure}
	\centering
	\includegraphics[width = \linewidth]{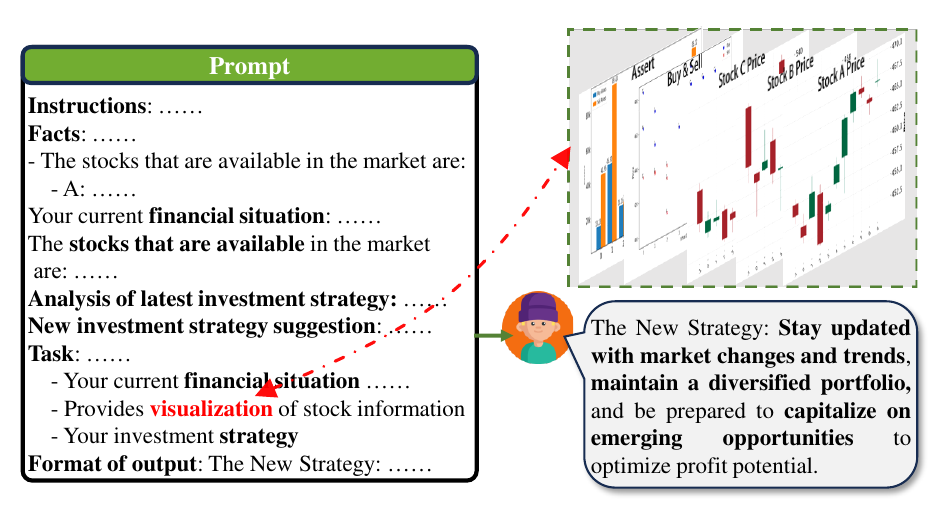}
	\caption{\textbf{Design of the Reflection Module.} 
	\textbf{Left:} Agent trading records, strategy evaluations, and top- and bottom-performing strategies from the strategy library. 
	\textbf{Top right:} Corresponding stock chart. 
	\textbf{Bottom right:} Newly generated strategy based on current information.}
	\label{fig:reflection}
	% \vspace{-4pt}
\end{figure}

% \vspace{-4pt}
\paragraph{Strategy Optimization via Reflection.} 

At the end of each trading day, the reflection module distills daily experiences into bidirectional learning signals (see \cref{fig:reflection}). It first evaluates the effectiveness of trading trajectories and strategies using a scoring function. Then, by contrasting top- and bottom-performing strategies, agents generate refined action plans and archive unsuccessful tactics in a strategy library. This iterative process drives continuous strategy evolution and more robust decision-making.

%More details can be found in \cref{appendix_arena}.

% \input{body/Experiment}

%To simulate a realistic stock market environment, we developed the \textit{Agent Trading Arena}, a controlled environment that isolates external factors. The system's general architecture is illustrated in \cref{fig: AgentArena}. In this environment, agents discuss stock market trends, analyze stock data, and engage in trading activities. Each agent is allowed to perform multiple trades daily and, at the end of each trading session, evaluate their performance and refine their strategies accordingly. 

\section{Experimental Results}

\subsection{Experimental Setup}

% \vspace{-4pt}
\paragraph{Environment.} 

The \textit{Agent Trading Arena} is lightweight and accessible, requiring only API credits and no GPU for deployment. Details on the selection of different LLMs are provided in \cref{appendix_LLM}.

\begin{table}
	\centering
	\footnotesize
	\setlength{\tabcolsep}{4pt}
	\begin{tabular}{cl|ccl}
	\toprule[1.1pt]
	ID & Name & \makecell{Duration \\ (years)} & \makecell{Initial \\ Capital} & Profession \\
	\midrule
	1 & Amy & 1 & 100,000 & AI Researcher \\
	2 & Bruce & 2 & 100,000 & Lawyer \\
	3 & Charles & 1 & 100,000 & Doctor \\
	4 & David & 3 & 100,000 & Engineer \\
	5 & Ella & 2 & 100,000 & Teacher \\
	6 & Frank & 5 & 100,000 & Entrepreneur \\
	7 & Grace & 4 & 100,000 & Accountant \\
	8 & Hank & 2 & 100,000 & Architect \\
	9 & Ivy & 3 & 100,000 & Marketing Manager \\
	\dots & \dots & \dots & \dots & \dots \\
	\bottomrule[1.1pt]
	\end{tabular}
	\caption{\textbf{Agent Characteristics.} \texttt{Duration} denotes each agent’s investment horizon, and \texttt{Initial Capital} indicates the starting funds for trading.}
	\label{tab:table_person}
\end{table}

\begin{table}
	\centering
	\footnotesize
	\setlength{\tabcolsep}{4pt}
	\begin{tabular}{cc|ccc}
	\toprule[1.1pt]
	ID & Ticker & DPS & Closing Prices & Initial Quantity \\
	\midrule
	1 & A & 22 & 454.17, \dots, 445.60 & 1,200 \\
	2 & B & 23 & 354.17, \dots, 465.80 & 1,000 \\
	3 & C & 25 & 500.47, \dots, 440.60 & 1,600 \\
	\dots & \dots & \dots & \dots & \dots \\
	\bottomrule[1.1pt]
	\end{tabular}
	\caption{\textbf{Stock Details.} \texttt{DPS} denotes dividend per share; \texttt{Closing Prices} are historical end-of-day closing prices; and \texttt{Initial Quantity} specifies the starting number of shares.}
	\label{tab:stock}
\end{table}

% \vspace{-4pt}
\paragraph{Datasets.}

We configure the Arena with varying numbers of agents and stocks, all initialized with the same capital (see \cref{tab:table_person,tab:stock}). To validate our agents, we simulate portfolio investments on subsets of \textsc{NASDAQ} and \textsc{CSI} datasets. For \textsc{NASDAQ}, we run a two-month simulation (September 3–October 29, 2024) with an initial capital of 100,000 units, excluding weekends and holidays. For \textsc{CSI}, we conduct a one-year simulation (January 2–December 31, 2024) with the same capital, also excluding non-trading days. Both datasets, collected from Yahoo Finance and Baostock, provide daily records of opening, closing, high, and low prices, trading volume, and associated technical indicators, offering a holistic and up-to-date depiction of market behavior. Additional details are given in \cref{appendix_NASDAQ}. 

% \vspace{-4pt}
\paragraph{Baselines.} 

We compare \texttt{ArenaTrader} against the following strategies: 

\textbf{1) Buy \& Hold}: A passive strategy that purchases assets initially and holds them throughout the period without further trading. 

\textbf{2) SMA}~\citep{sma}: A moving average crossover strategy used to detect bullish and bearish momentum. We evaluate window sizes [5, 10, 15, 20] and select the best based on validation results. 

\textbf{3) ZMR}~\citep{zmr}: A zone-based reversal strategy that trades on deviations from preset bounds, with thresholds and holding periods tuned via validation. 

\textbf{4) MACD}~\citep{macd}: A momentum strategy based on MACD and signal line crossovers, where MACD is the 12-day EMA minus the 26-day EMA, and the signal line is the 9-day EMA. 

\textbf{5) StockFormer}~\citep{stockformer}: A hybrid model combining predictive coding and reinforcement learning to capture future dynamics and asset correlations. We follow the default setup from the original implementation. 

\textbf{6) TimesNet}~\citep{timesnet}: Transforms 1D time series into 2D representations to model multi-period temporal variations. We follow the original setup and select the look-back window from [5, 10, 15, 20] based on validation. 

\textbf{7) StockMixer}~\citep{stockmixer}: A lightweight MLP that mixes indicators, time, and stocks for efficient modeling of market dynamics. We follow the settings specified in the original work. 

% \vspace{-4pt}
\paragraph{Metrics.} 

We evaluate using five metrics: 

\textbf{1) Total Return (TR)}: $(C_1 - C_0)/C_0$, where $C_0$ and $C_1$ are initial and final capital. 

\textbf{2) Win Rate (WR)}: $N_w/N_t$, the ratio of profitable trading days to total trading days. 

\textbf{3) Sharpe Ratio (SR)}: $(R_p - R_f)/\sigma_p$, where $R_p$ is the mean daily return, $\sigma_p$ is the standard deviation of daily returns, and $R_f$ is set to 0 to emphasize risk-adjusted returns. 

\textbf{4) Mean Daily Return (Mean)}: Average daily return over the simulation. 

\textbf{5) Return Volatility (Std)}: Standard deviation of daily returns, measuring risk. 

\subsection{Experimental Results}

In the \textit{Agent Trading Arena}, we configure multiple agents and LLM backbones to promote diversity. The primary agent, \texttt{ArenaTrader}, consistently uses our full-featured design, including visual perception and reflection, enabling focused analysis. We evaluate \texttt{ArenaTrader} on real-world datasets, with corresponding prompts detailed in \cref{appendix_prompt}.

\begin{table*}
	\centering
	\footnotesize
	\setlength{\tabcolsep}{5pt}
	\begin{tabular}{l|ccccccc}
	\toprule[1.1pt]
	Model & Avg. Trend & TR (\%) $\uparrow$ & Mean (\%) $\uparrow$ & Std (\%) $\downarrow$ & WR (\%) $\uparrow$ & SR $\uparrow$ & $\Delta$ (\%) $\uparrow$ \\
	\midrule
	\texttt{ArenaTrader} \textit{w/} Qwen-VL-128k & -0.4632 & 33.3018 & 2.6526 & 1.0150 & \textbf{100.00} & 2.6135 & 33.7650 \\
	\texttt{ArenaTrader} \textit{w/} GPT-4o-mini & 5.4722 & 38.6460 & 3.0249 & 1.4235 & \textbf{100.00} & 2.1249 & 33.1738 \\
	\texttt{ArenaTrader} \textit{w/} Gemini-1.5 & 7.2743 & \underline{41.3264} & \underline{3.1946} & \textbf{0.1588} & \textbf{100.00} & \textbf{20.1128} & 34.0521 \\
	\texttt{ArenaTrader} \textit{w/} Qwen-VL-32k & -0.5216 & 35.8150 & 3.1213 & 1.6079 & 90.00 & 1.9412 & \underline{36.3366} \\
	\rowcolor{gray!20}
	\texttt{ArenaTrader} \textit{w/} GPT-4o & 3.5881 & \textbf{47.6851} & \textbf{3.6095} & \underline{0.5327} & \textbf{100.00} & \underline{6.7765} & \textbf{44.0970} \\
	\bottomrule[1.1pt]
	\end{tabular}
	\caption{\textbf{Investment Performance of \texttt{ArenaTrader} with Different LLMs.} \texttt{Avg. Trend} denotes the average market trend, and \texttt{$\Delta$} indicates the gain relative to this trend. GPT-4o achieves the highest returns and stability, demonstrating superior data-driven reasoning among vision-capable LLMs. Bold and underlined values denote the best and second-best results, respectively.}
	\label{tab:LLM_vl}
\end{table*}

\begin{figure}
	\centering
	\tabcolsep = 1pt
	\begin{tabular}{cc}
	\includegraphics[width = 0.48\linewidth]{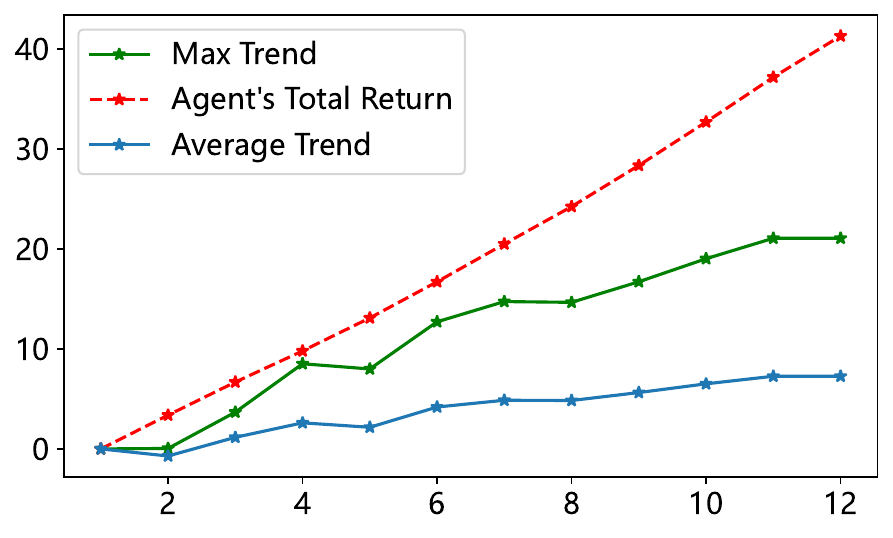} & 
	\includegraphics[width = 0.48\linewidth]{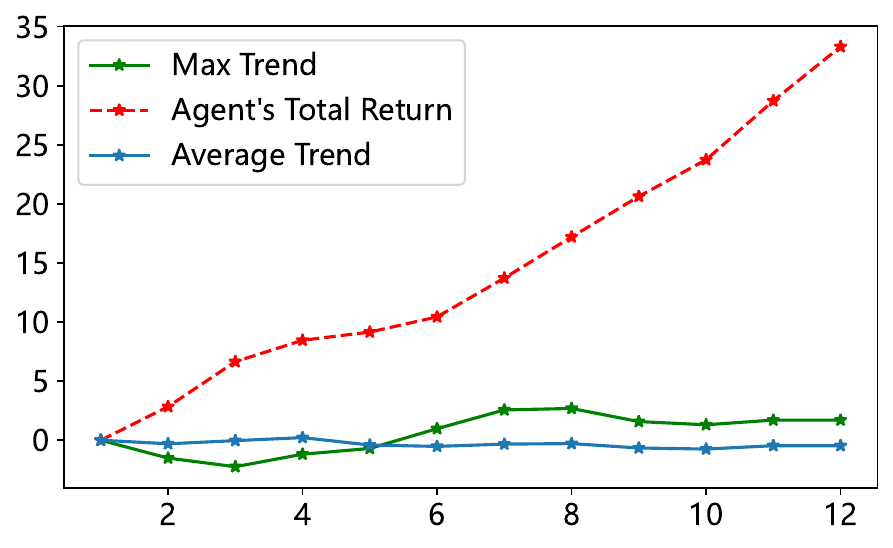} 
	\vspace{-4pt} \\
	\footnotesize{(a) Gemini-1.5} & \footnotesize{(b) Qwen-VL-128k} \\
	\includegraphics[width = 0.48\linewidth]{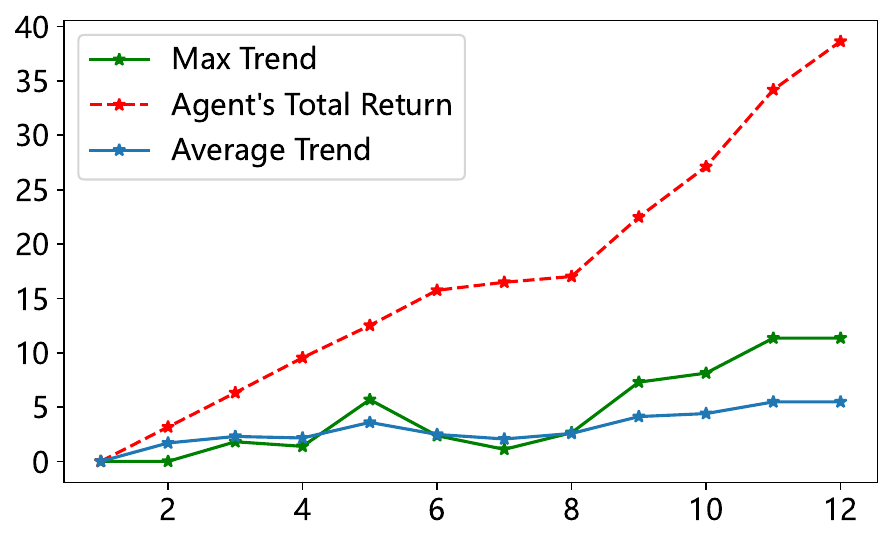} & 
	\includegraphics[width = 0.48\linewidth]{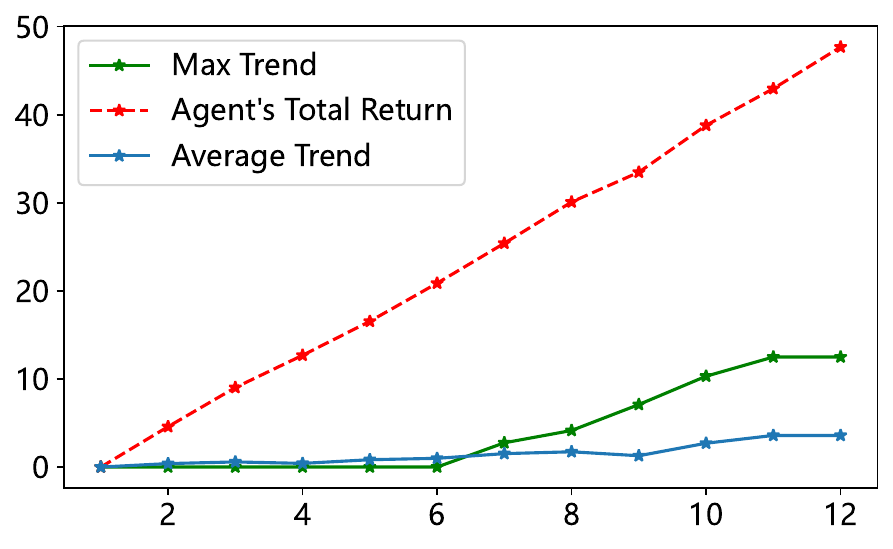}
	\vspace{-4pt} \\
	\footnotesize{(c) GPT-4o-mini} & \footnotesize{(d) GPT-4o} 
	\end{tabular}
	\caption{\textbf{\texttt{ArenaTrader} Performance \textit{vs.} Market Volatility.} Total return of \texttt{ArenaTrader} in the \textit{Agent Trading Arena}, compared with the most volatile single stock and the average return across all stocks. This comparison highlights the agent’s robustness under varying market conditions.}
	\label{fig:four}
	% \vspace{-4pt}
\end{figure}

% \vspace{-4pt}
\paragraph{Vision-Enhanced Agent Performance.} 

\cref{tab:LLM_vl,fig:four} report the returns of \texttt{ArenaTrader} under different vision-capable LLMs. GPT-4o~\citep{GPT4o} achieves the highest TR, followed by Gemini-1.5~\citep{Gemini1.5}. For volatility-adjusted returns, GPT-4o again leads, with Qwen-VL-32k~\citep{Qwen2VL} second. These results highlight GPT-4o’s strong data-driven reasoning and robustness across varying volatility regimes, underscoring its ability to integrate visual and numerical signals in dynamic markets. 

%\vspace{-3pt}

%\begin{table}
%	\centering
%	\footnotesize
%	\setlength{\tabcolsep}{5pt}
%	\begin{tabular}{lrc}
%	\toprule[1.1pt]
%	Strategy & \multicolumn{1}{c}{TR $\uparrow$} & SR $\uparrow$ \\
%	\midrule
%	MACD~\citep{macd} & 7.18 & 0.173 \\
%	StockFormer~\citep{stockformer} & 9.05 & 0.073 \\
%	TimesNet~\citep{timesnet} & 11.74 & 0.203 \\
%	StockMixer~\citep{stockmixer} & 14.64 & 0.305 \\
%	\midrule
%	GPT-4o + Textual & 8.69 & 0.167 \\
%	GPT-4o + Visual & 9.91 & 0.195 \\
%	GPT-4o + Textual + Visual & 12.59 & 0.253 \\
%	GPT-4o + Textual + Visual + Reflection & \textbf{15.99} & \textbf{0.313} \\
%	\bottomrule[1.1pt]
%	\end{tabular}
%	\caption{\textbf{Reproduced Backtesting Performance on \textsc{NASDAQ Stock} Dataset.} The agent with textual and visual input outperformed the NASDAQ-100 by \textbf{53.97\%} in SR during the same period.}
%	\label{real_data}
	%\vspace{-3pt}
%\end{table}

\begin{table*}
	\centering
	\footnotesize
	\setlength{\tabcolsep}{5pt}
	\begin{tabular}{l|ccccc|ccccc}
	\toprule[1.1pt]
	\multirow{2}[2]{*}{Strategy} 
	& \multicolumn{5}{c|}{\textsc{NASDAQ}} 
	& \multicolumn{5}{c}{\textsc{CSI}} \\
	\cmidrule(lr){2-6} \cmidrule(lr){7-11}
	& TR $\uparrow$ & Mean $\uparrow$ & Std $\downarrow$ & WR $\uparrow$ & SR $\uparrow$
	& TR $\uparrow$ & Mean $\uparrow$ & Std $\downarrow$ & WR $\uparrow$ & SR $\uparrow$ \\
	\midrule
	Buy \& Hold & 6.06 & 0.155 & 0.916 & \textbf{61.54} & 0.169 & 0.73 & 0.004 & 0.403 & 43.15 & 0.010 \\
	SMA & 0.19 & 0.006 & \underline{0.561} & 46.15 & 0.011 & -3.45 & -0.014 & \textbf{0.319} & 33.20 & -0.044 \\
	ZMR & 1.04 & 0.030 & 0.814 & 41.03 & 0.037 & 1.24 & 0.006 & \underline{0.361} & 41.91 & 0.016 \\
	MACD & 5.62 & 0.146 & 1.033 & 58.97 & 0.141 & -0.67 & -0.002 & 0.373 & 40.66 & -0.006 \\
	\midrule
	StockFormer~\citep{stockformer} & 11.23 & 0.281 & \textbf{0.121} & 56.41 & 0.231 & 14.83 & \textbf{0.636} & 1.128 & 41.08 & 0.056 \\
	TimesNet~\citep{timesnet} & 11.41 & 0.283 & 1.035 & 56.41 & 0.273 & 32.93 & 0.139 & 2.053 & \underline{48.55} & 0.068 \\
	StockMixer~\citep{stockmixer} & \underline{14.64} & \underline{0.357} & 1.173 & 56.41 & \underline{0.305} & \underline{43.00} & 0.153 & 0.907 & 27.80 & \textbf{0.168} \\
	\rowcolor{gray!20}
	\texttt{ArenaTrader} \textit{w/} GPT-4o (Ours) & \textbf{15.99} & \textbf{0.450} & 1.110 & \underline{58.97} & \textbf{0.348} & \textbf{48.93} & \underline{0.176} & 1.432 & \textbf{51.45} & \underline{0.123} \\
	\bottomrule[1.1pt]
	\end{tabular}
	\caption{\textbf{Strategy Performance on \textsc{NASDAQ} and \textsc{CSI}.} Our method demonstrates strong robustness and generalization, achieving near-optimal performance without additional training.}
	% Bold and underlined values denote the best and second-best results, respectively.}
	\label{tab:nasdaq_ashare}
\end{table*}

% \vspace{-4pt}
\paragraph{Evaluation on Real-World Data.} 

Using GPT-4o as the backbone, \cref{tab:nasdaq_ashare} presents results on \textsc{NASDAQ} and \textsc{CSI}. Training details for StockFormer~\citep{stockformer}, TimesNet~\citep{timesnet}, and StockMixer~\citep{stockmixer} are provided in \cref{appendix_NASDAQ}. Despite extensive training, these models underperform our method, which uses a 10-day input window selected via ablation. The Sharpe Ratio (SR) of \textsc{NASDAQ}-100 benchmark during the same period is 0.189, while \textsc{CSI} 300 records 0.054. In contrast, our model achieves 0.348 on \textsc{NASDAQ}-100 (an improvement of 0.159) and 0.123 on \textsc{CSI} 300 (an improvement of 0.069). 

\begin{figure}
	\centering
	\begin{tabular}{c}
	\includegraphics[width = 0.92\linewidth]{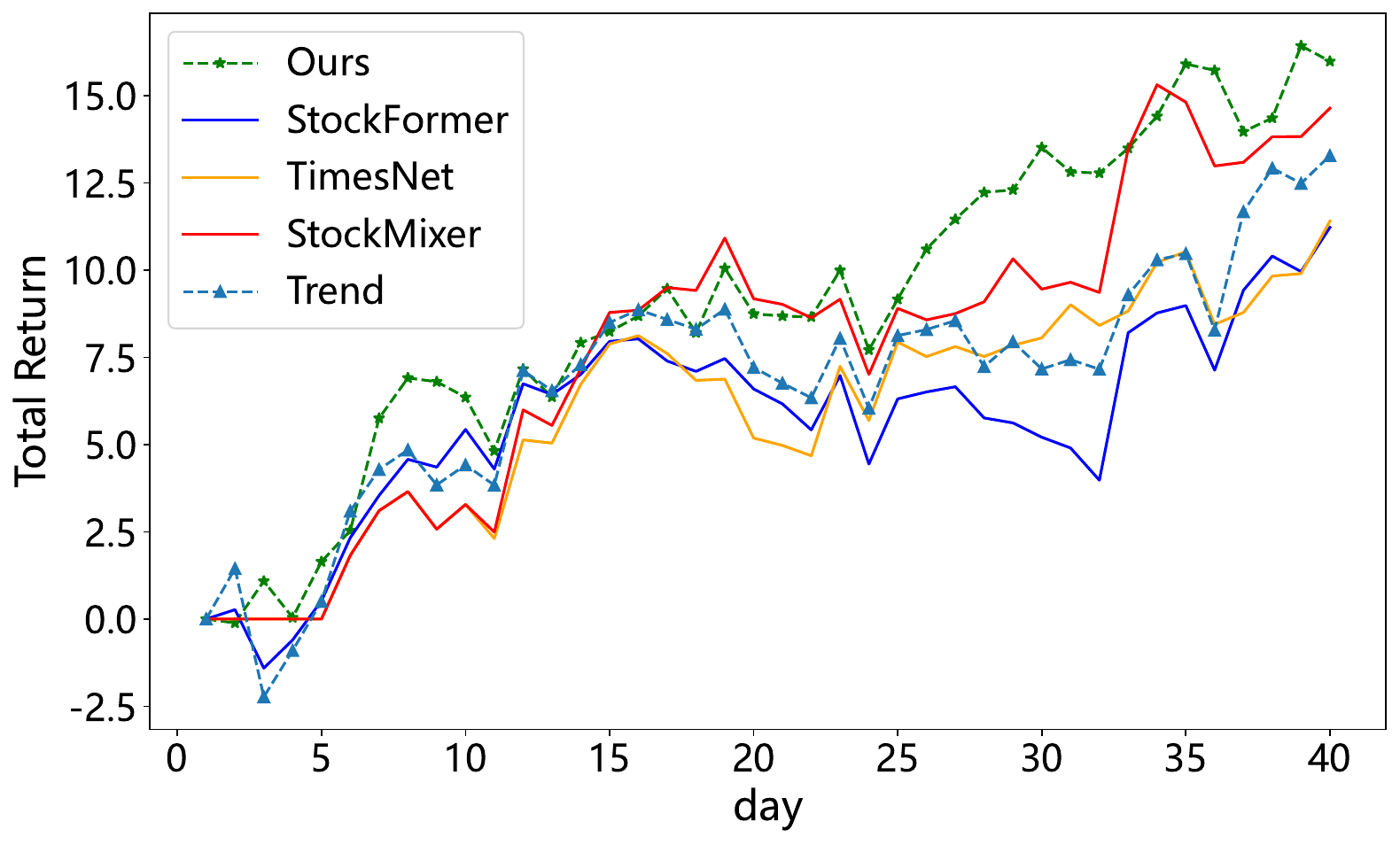} \vspace{-4pt} \\
	\footnotesize{(a) \textsc{NASDAQ}} \\
	\includegraphics[width = 0.92\linewidth]{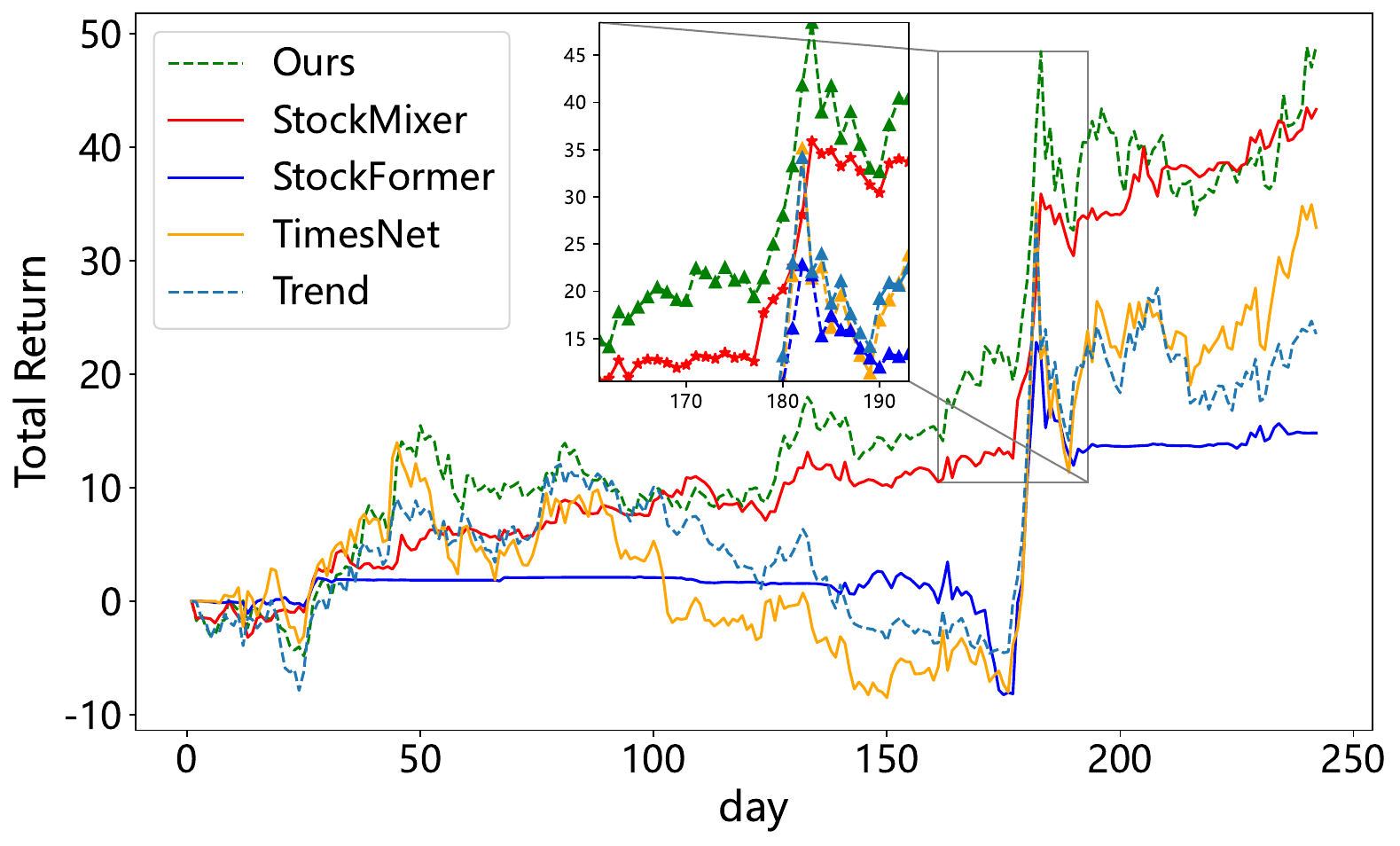} \vspace{-4pt} \\
	\footnotesize{(b) \textsc{CSI}} 
	\end{tabular}
	\caption{\textbf{Backtesting Results on Real-World Stock Datasets.} Two-month backtest on \textsc{NASDAQ} and one-year backtest on \textsc{CSI}. The proposed method consistently delivers near-optimal performance and outperforms all baselines, particularly during periods of high market volatility.}
	\label{fig:combined_returns}
	% \vspace{-4pt}
\end{figure}

Building on these results, \cref{fig:combined_returns} provides a detailed analysis of portfolio trajectories and market conditions, illustrating our agent’s near-optimal performance, especially during high-volatility periods. Notably, our approach operates without task-specific retraining, relying solely on historical price data. This design underscores the robustness and strong generalization ability of our agent in real-world, dynamic market environments.

% Our agent requires no additional training, relying solely on historical prices for decision-making, which highlights its robustness and generalization in dynamic markets. 

% The Sharpe ratios for Apple, NASDAQ-100, and S\ & P 500 during the same period were 0.097, 0.189, and 0.205, respectively. The agent with textual and visual input outperformed the NASDAQ-100 and S\ & P 500 in Sharpe Ratio by \textbf{53.97\%} and \textbf{41.95\%}, respectively. Moreover, only visual input outperformed textual input, further highlighting LLMs' stronger geometric reasoning abilities.

\subsection{Ablation Studies}

We conduct ablation studies in the Arena to evaluate the impact of input formats and the reflection mechanism on agent performance. To further assess the effect of sequence length across textual and visual modalities, we perform controlled backtesting on \textsc{NASDAQ}. Backtesting mitigates dynamic data fluctuations, allowing sequence length to be isolated as the primary variable. 

\begin{table*}
	\centering
	\footnotesize
	\setlength{\tabcolsep}{9pt}
	\begin{tabular}{c|l|cc|ccccc}
	\toprule[1.1pt]
	& Model & Textual & Visual 
	& TR $\uparrow$ & Mean $\uparrow$ & Std $\downarrow$ 
	& WR $\uparrow$ & SR $\uparrow$ \\
	\midrule
	\multirow{9}[4]{*}{\rotatebox{90}{\textit{w/o} Reflection}} 
	& LLaMa-3~\citep{LLaMa3} & \CIRCLE & \Circle & 4.09 & 0.686 & 2.270 & 62.50 & 0.196 \\
	& DeepSeek~\citep{DeepSeek3v} & \CIRCLE & \Circle & 27.31 & 2.248 & 2.755 & 92.21 & 1.530 \\
	& Qwen-2.5~\citep{qwen2.5} & \CIRCLE & \Circle & 30.37 & 2.680 & 1.459 & 93.75 & 1.703 \\
	\cmidrule(l){2-9}
	& \multirow{3}{*}{Gemini-1.5~\citep{Gemini1.5}} & \CIRCLE & \Circle & 14.32 & 1.221 & 2.778 & 83.84 & 1.729 \\
	& & \Circle & \CIRCLE & 19.04 & 1.604 & 1.165 & 90.91 & 1.376 \\
	& & \CIRCLE & \CIRCLE & 23.96 & 1.981 & 1.334 & \textbf{100.00} & 1.485 \\
	\cmidrule(l){2-9}
	& \multirow{3}{*}{GPT-4o~\citep{GPT4o}} & \CIRCLE & \Circle & 13.04 & 1.139 & 1.929 & 54.55 & 0.590 \\
	& & \Circle & \CIRCLE & 17.07 & 1.454 & 1.504 & 72.73 & 0.967 \\
	& & \CIRCLE & \CIRCLE & 26.18 & 2.157 & 2.058 & 90.91 & 1.048 \\
	\midrule
	\multirow{9}[4]{*}{\rotatebox{90}{\textit{w/} Reflection}} 
	& LLaMa-3~\citep{LLaMa3} & \CIRCLE & \Circle & 10.25 & 1.667 & 2.403 & 66.67 & 0.694 \\
	& DeepSeek~\citep{DeepSeek3v} & \CIRCLE & \Circle & 30.62 & 2.470 & 1.512 & \textbf{100.00} & 1.633 \\
	& Qwen-2.5~\citep{qwen2.5} & \CIRCLE & \Circle & 38.91 & \underline{3.324} & 1.073 & \textbf{100.00} & 3.099 \\
	\cmidrule(l){2-9}
	& \multirow{3}{*}{Gemini-1.5~\citep{Gemini1.5}} & \CIRCLE & \Circle & 29.45 & 2.519 & 3.012 & 93.51 & 2.210 \\
	& & \Circle & \CIRCLE & 37.01 & 2.911 & 1.235 & \textbf{100.00} & 2.357 \\
	& & \CIRCLE & \CIRCLE & \underline{41.33} & 3.195 & \textbf{0.159} & \textbf{100.00} & \textbf{20.113} \\
	\cmidrule(l){2-9}
	& \multirow{3}{*}{GPT-4o~\citep{GPT4o}} & \CIRCLE & \Circle & 33.65 & 2.671 & 2.171 & 98.70 & 2.142 \\
	& & \Circle & \CIRCLE & 35.76 & 2.821 & 0.678 & \textbf{100.00} & 4.159 \\
	& & \cellcolor{gray!20}\CIRCLE & \cellcolor{gray!20}\CIRCLE &\cellcolor{gray!20}\textbf{47.69} & \cellcolor{gray!20}\textbf{3.610} & \cellcolor{gray!20}\underline{0.533} & \cellcolor{gray!20}\textbf{100.00} & \cellcolor{gray!20}\underline{6.777} \\
	\bottomrule[1.1pt]
	\end{tabular}
	\caption{\textbf{Performance Comparison with and without Reflection Using Textual and Visual Inputs.} Agents using visual data consistently outperform those using text alone, while combining both modalities yields the strongest results.} 
	% Bold and underlined values denote the best and second-best results, respectively.}
	\label{tab:LLMS_9agent_reflect}
\end{table*}

% \vspace{-4pt} 
\paragraph{Textual \textit{vs.} Visual Inputs.} 

We compare agents using textual-only inputs with those leveraging visualizations (scatter plots, line charts, bar graphs). Non-vision LLMs receive text only without reflection, while vision-capable LLMs are evaluated under text-only, visual-only, and combined-input settings. Corresponding prompts are provided in \cref{appendix_prompt}, and details of LLM choices are in \cref{appendix_LLM}. Results in \cref{tab:LLMS_9agent_reflect} show that visual inputs substantially improve returns, with combined inputs achieving the best overall performance.

% \vspace{-4pt}
\paragraph{Impact of Reflection.} 

Incorporating the reflection module further enhances performance across all LLMs. For GPT-4o, TR improves from 33.65\% with text-only to 47.70\% with visual input, a relative gain of 41.7\%. Beyond TR, reflection also boosts SR and WR, indicating higher profitability, stability, and consistency in trading outcomes. These findings confirm that strategic reflection, particularly with visual data, strengthens agents’ reasoning and decision-making under volatile market conditions. 

\begin{table}[t]
	\centering
	\footnotesize
	\setlength{\tabcolsep}{3pt}
	\begin{tabular}{c|c|ccccc}
	\toprule[1.1pt]
	Input & Window & TR $\uparrow$ & Mean $\uparrow$ & Std $\downarrow$ & WR $\uparrow$ & SR $\uparrow$ \\
	\midrule
	\multirow{4}{*}{Textual} 
	& 5 days & 3.63 & 0.102 & 1.588 & 62.5 & 0.064 \\
	& 10 days & 5.58 & 0.145 & 1.334 & 60.0 & 0.109 \\
	& 15 days & 6.47 & 0.171 & 1.681 & 57.5 & 0.102 \\
	& 20 days & 1.67 & 0.050 & 1.296 & 62.5 & 0.039 \\
	\midrule
	\multirow{4}{*}{Visual} 
	& 5 days & 8.21 & 0.207 & 1.396 & 57.5 & 0.149 \\
	& 10 days & 6.90 & 0.174 & 1.164 & \textbf{67.5} & 0.149 \\
	& 15 days & 6.90 & 0.174 & 1.216 & 60.0 & 0.143 \\
	& 20 days & 2.81 & 0.075 & \textbf{1.101} & 62.5 & 0.068 \\
	\midrule
	\multirow{4}{*}{\makecell{Textual \\ + \\ Visual}} 
	& 5 days & 14.33 & 0.372 & 1.719 & 60.0 & 0.216 \\
	& \cellcolor{gray!20}10 days& \cellcolor{gray!20}\textbf{15.99} & \cellcolor{gray!20}\textbf{0.450} & \cellcolor{gray!20}\underline{1.110} & \cellcolor{gray!20}59.0 & \cellcolor{gray!20}\textbf{0.348} \\
	& 15 days & \underline{14.73} & \underline{0.380} & 1.684 & 57.5 & \underline{0.226} \\
	& 20 days & 13.51 & 0.352 & 1.577 & \underline{65.0} & 0.223 \\
	\bottomrule[1.1pt]
	\end{tabular}
	\caption{\textbf{Performance with Textual and Visual Inputs Across Different Window Sizes.} \texttt{Window} denotes the number of consecutive trading days provided as historical input before the current trading day. \texttt{Textual + Visual} consistently achieves the best performance, particularly with a 10-day window, highlighting the importance of both modality choice and time-window design.} 
	% Bold and underlined values denote the best and second-best results, respectively.}
 \label{tab:modality_ablation}
\end{table}

% \vspace{-4pt}
\paragraph{Impact of Time-Series Length.} 

Experiments with GPT-4o as the backbone are summarized in \cref{tab:modality_ablation}, which reports performance across modalities and time windows. Overall, ``Textual + Visual'' consistently outperforms single modalities, particularly with a 10-day window, achieving the highest TR (15.99\%), SR (0.348), and Mean (0.450\%). Visual inputs perform best in the 5–10 day range, offering strong returns, stability, and win rates. Performance declines at 20 days, likely due to noise accumulation or overfitting. These results underscore the importance of both multi-modal fusion and careful time-window selection for robust time-series decision-making.

\section{Conclusion}

We presented the \textit{Agent Trading Arena}, a closed-loop, zero-sum simulation framework designed to model complex market dynamics and evaluate LLMs on numerical reasoning tasks. Our experiments show that while LLMs struggle with plain-text numerical data, they achieve significantly higher returns when supported by visual representations, underscoring the importance of charts and graphs in complex decision-making. Integrating a reflection module further improves performance by enabling iterative analysis and strategic refinement. Validation on both \textsc{NASDAQ} and \textsc{CSI} confirms that vision-enhanced LLM agents consistently outperform text-only counterparts without additional training. Overall, this work highlights both the capabilities and limitations of LLMs in numerical reasoning and establishes the \textit{Agent Trading Arena} as a robust testbed for advancing research in adaptive, real-world financial applications.

\section*{Limitations}

This study investigates the visual reasoning abilities of LLMs in a virtual stock trading environment. While such a controlled setup enables systematic analysis, it inevitably simplifies many aspects of real-world financial systems, including uncertainty, feedback loops, and human interactions. Bridging the gap between virtual environments and real-world applications therefore remains a significant challenge. 

A promising direction for future work is to explore how agents can be gradually integrated into real financial workflows, interacting with live data, adapting to real-time signals, and contributing meaningful decisions or strategies. This would move beyond passive evaluation toward active participation, where agents not only respond to environments but also generate novel behavioral data and insights to inform downstream analysis, model training, and system design. Expanding the task scope and incorporating more diverse modalities may further support the development of generalizable and robust evaluation frameworks.

\section*{Acknowledgments}

This work was supported in part by the National Natural Science Foundation of China under Grants 62301213 and 62271361, the Hubei Provincial Key Research and Development Program under Grant 2024BAB039, and the Open Project Funding of the Hubei Key Laboratory of Big Data Intelligent Analysis and Application, Hubei University, under Grant 2024BDIAA01.

\bibliography{SQIAgent}

\clearpage

\appendix

\section{Agent Trading Arena}
\label{appendix_arena}

\subsection{Overview}

\begin{figure*}
	\centering
	\includegraphics[width = \linewidth]{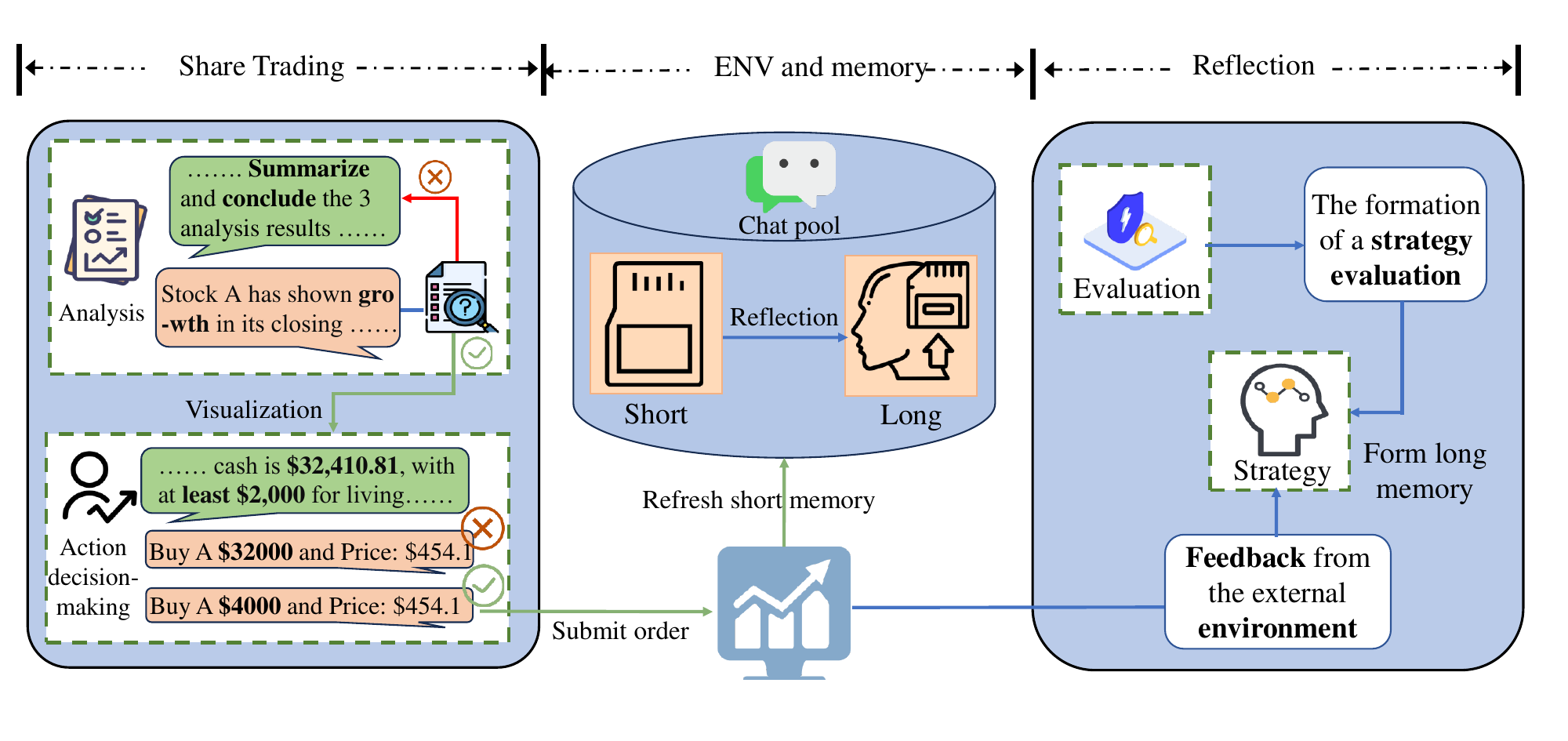}
	\caption{\textbf{Agent Workflow Components:} Share Trading involves stock analysis, decision validation, and trade execution; Environment and Memory manage memory and process trade orders; and Reflection focuses on strategy assessment and refinement based on feedback.}
	\label{Agent}
	% % \vspace{-4pt}
\end{figure*}

As illustrated in \cref{Agent}, each agent's workflow integrates LLMs for chat pool interactions, stock analysis, decision-making, and reflection. In the stock analysis and decision-making modules, all outputs are validated for consistency with both common sense and operational requirements before execution. 

\subsubsection{Action Decision-Making}

\begin{figure}[t]
	\centering
	\includegraphics[width = \linewidth]{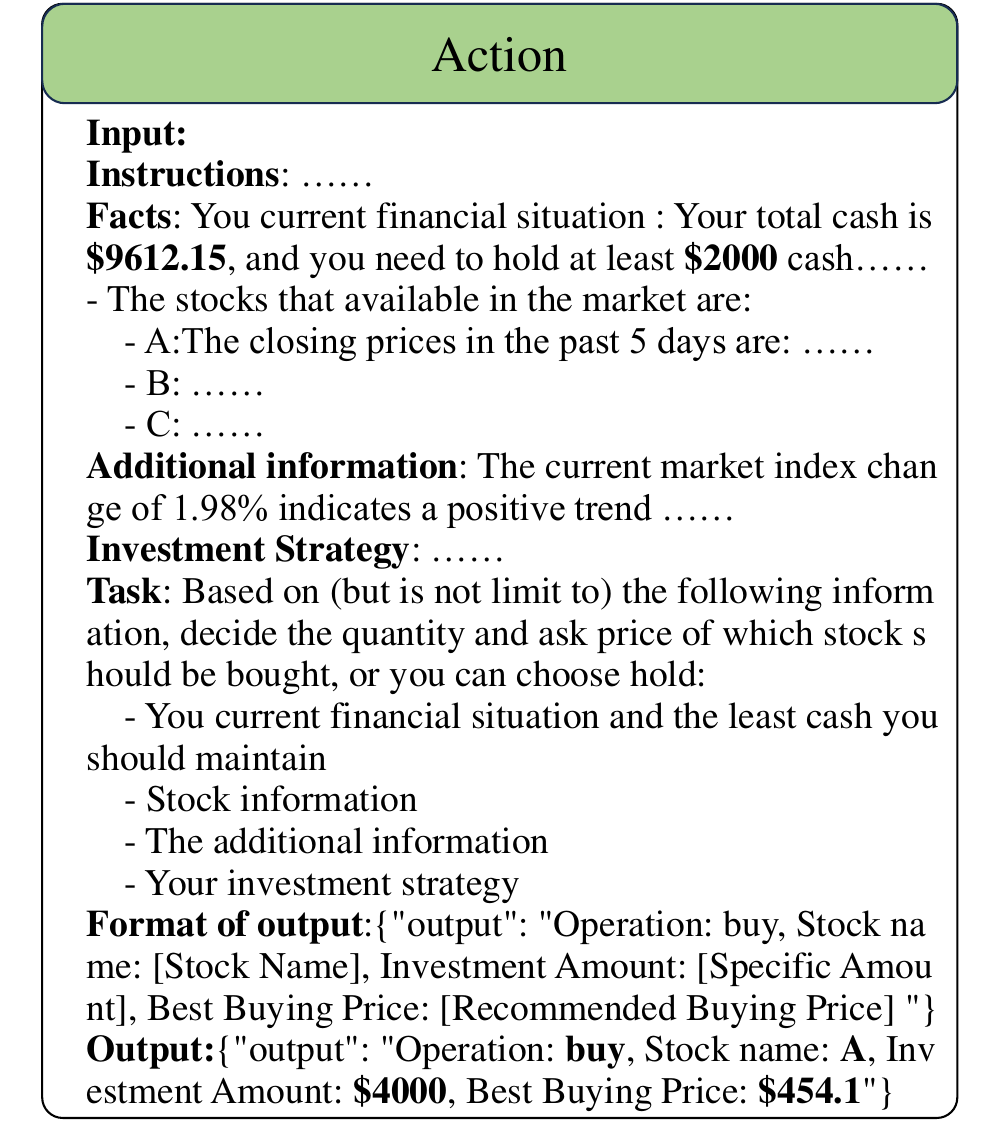} 
	\caption{\textbf{Inputs and Outputs in Action Decision-Making.}}
 	\label{action}
	% % \vspace{-4pt}
\end{figure}

Action generation follows the LLM framework. The agent responsible for generating actions receives corresponding prompts via SQLite. Based on these prompts and specified output formats, the agent decides whether to buy, sell, or hold stocks. The action generation process is outlined in \cref{Action}, with input prompts shown in \cref{action} and the corresponding outputs displayed in the adjacent figure.
\begin{equation}\label{Action}
	\begin{cases} 
	A_{\mathrm{date} + 1}^t = \Psi \left(\mathrm{Ins}, Z_\mathrm{date}^t, S_\mathrm{date} \right), & \text{if~} t \text{~is Iters}, \\
	A_\mathrm{date}^{t + 1} = \Psi \left(\mathrm{Ins}, Z_\mathrm{date}^t, S_\mathrm{date} \right), & \text{otherwise},
	\end{cases}
\end{equation}
where $\mathrm{Ins}$ represents the environment introduction, $Z_\mathrm{date}^t$ denotes the memory of the stock transaction on day $\mathrm{date}$ retrieved from the database, and $S_\mathrm{date}$ is the strategy for day $\mathrm{date}$ generated via reflection. 

\subsubsection{Environmental Interaction}
\label{apppendix_Enxir_inteusedraction}

To isolate external influences, we created a virtual sandbox environment where each agent is assigned a unique ID, and their actions affect the environment. The function $\phi$ facilitates environmental interactions, as shown in \cref{Algorithm_env}, where ``OPS'' retrieves agent actions, ``date'' refers to the trading date, and ``Z'' represents the memory used for interaction with the environment. Through $\phi$, each agent's actions, such as buying or selling stocks, determine the current stock price and update the trading platform, including stock prices and available shares. Stock prices are independent of external factors and are influenced solely by the sandbox's internal dynamics. The stock price is updated with each transaction according to the following formula:
\begin{equation}\label{price_update}
 	\begin{aligned}
 	\mathrm{Price}_\mathrm{curr} & = \delta \left(Q, F, \mathrm{Price}_\mathrm{curr}, \mathrm{Price}_\mathrm{deal} \right) \\
	& = \frac{\mathrm{Price}_\mathrm{deal} \cdot Q \cdot F + \mathrm{Price}_\mathrm{curr} \cdot Q_\mathrm{total}}{Q \cdot F + Q_\mathrm{total}},
 	\end{aligned} 
\end{equation}
where $Q$ is the quantity of stock traded, $F$ is the fluctuation constant, $\mathrm{Price}_\mathrm{curr}$ is the current stock price, $Q_\mathrm{total}$ is the total number of shares available, and $\mathrm{Price}_\mathrm{deal}$ is the price at which the trade occurs. 

\SetAlgorithmName{Algorithm}{Algorithm}{List of Algorithms}
\SetAlFnt{\small}
\SetKwInput{KwIn}{Input}
\SetKwInput{KwOut}{Output}

\begin{algorithm}
\caption{\small Environmental Interaction}
\label{Algorithm_env}
\KwIn{OPS: Function to retrieve agents' actions, 
 date: The current trading date, 
 Z: Memory used for interaction with the environment}
\KwOut{Z}
\For{$P \in \text{Persons}$}{
 	$A \leftarrow \text{OPS}(t, P)$\;
 	$O, N, Q, \text{Price}_{\text{deal}} \leftarrow \text{Extract}(A)$\;
 	$\text{Price}_{\text{curr}}, Q_{\text{total}} \leftarrow \text{Stocks}(N)$\;
 	$\text{Price}_{\text{curr}} \leftarrow \delta(Q, F, \text{Price}_{\text{curr}}, \text{Price}_{\text{deal}})$\;
 	
 	\If{$O = $ ``buy''}{
 	$\text{Cash} \leftarrow \text{Price}_{\text{curr}} \cdot Q$\;
 	\If{$\text{Cash} < P.\text{Cash}$}{
 	$Z \leftarrow \text{SubmitOrder}(O, N, t, \text{Price}_{\text{curr}}, Q)$\;
 	}
 	}
 	
 	\If{$O = $ ``sell''}{
 	$\text{Hold} \leftarrow P(N)$\;
 	\If{$\text{Hold} \neq \text{None}$}{
 	$Q_N \leftarrow \text{Hold}[\text{``Q''}] - Q$\;
 	\If{$Q_N > 0$}{
 	$Z \leftarrow \text{SubmitOrder}(O, N, t, \text{Price}_{\text{curr}}, Q)$\;
 	}
 	}
 	}
 	}

$\text{Market}(\text{date}, \text{Persons})$\;
\Return Z\;
\end{algorithm}

The function $\phi$ executes trading orders and updates the stock price in real-time based on the agent's actions. To prevent excessive volatility and mitigate risk during trading cycles, a daily price fluctuation cap is enforced. Before executing any transaction, each agent evaluates its available funds and refrains from proceeding if insufficient capital is available.

\subsubsection{Memory} 

The superior performance of LLM-based agents arises from the extensive internal knowledge acquired during pre-training. The large number of parameters in LLMs enables the retrieval of diverse information and supports logical and inferential reasoning. To further enhance knowledge retrieval across various tasks, we incorporate a memory module that empowers LLM-based agents with self-improvement capabilities. This memory module facilitates strategy reflection through time-series feedback. Unlike qualitative tasks, quantitative feedback evolves incrementally with subtle differences, presenting a challenge for the generalization of existing LLM-based agents.

To minimize the influence of pre-existing knowledge, we assign specific roles to each agent. The memory module records sensory inputs and accumulates valuable experiences based on immediate feedback following actions. These experiences are stored in a database for future reference. The interaction history between an agent's behavior and the environment constitutes short-term memory, enabling the agent to retain recent events. The historical trajectory is defined as:
\begin{equation}\label{Eq1}
 	M_\mathrm{date}^t = \zeta \left(I_\mathrm{date}^t, \mathrm{Out}_\mathrm{date}^t \right),
\end{equation}
where $\zeta$ processes key information, $I_\mathrm{date}^t$ represents the input prompt, and $\mathrm{Out}_\mathrm{date}^t$ denotes the resulting output.

The day's trajectory constitutes short-term memory, which is expressed as:
\begin{equation}\label{Eq2}
 	Z_\mathrm{date}^t = \left(M_\mathrm{date}^0, M_\mathrm{date}^1, \dots, M_\mathrm{date}^t\right).
\end{equation}

The process for updating short-term memory is given by:
\begin{equation}\label{Eq3}
 	Z_\mathrm{date}^{t + 1} = M_\mathrm{date}^t \cup Z_\mathrm{date}^t, \quad t \in \{0,1,\dots,T\},
\end{equation}
where $Z_\mathrm{date}^t$ represents the short-term memory, and $T$ denotes the maximum number of iterations.

The reflection model serves as long-term memory, enabling self-reflection and the consolidation of knowledge.

\subsubsection{Reflection}
\label{appendix_reflection}

We propose a strategy distillation method that transforms quantitative results into descriptive text, which is then used as prompts for LLMs. This approach aids in the analysis of results and the generation of actionable, qualitative summaries, enabling LLMs to derive new strategies. These strategies are implemented, monitored, and evaluated over time, while underperforming strategies are archived for future review.

Initially, we evaluate the day’s trajectory memory and associated strategies. The evaluation function is defined as:
\begin{equation}\label{evaluation}
	E_\mathrm{date} = \Psi \left(\mathrm{Ins}, Z_\mathrm{date}^T, S_\mathrm{date} \right),
\end{equation}
where $\Psi$ represents the evaluation function based on LLMs, $\mathrm{Ins}$ contains the agent role descriptions and output requirements, and $Z_\mathrm{date}^T$ denotes the memory for the given day.

Based on the evaluation results and previous strategies, we generate the latest strategy. Past strategies are stored in a library and scored. For the new strategy, we select the top five best-performing and the bottom five worst-performing strategies to provide both positive and negative feedback. The strategy update formula is:
\begin{equation}\label{reflection}
	S_{\mathrm{date} + 1} = \Psi \left(\mathrm{Ins}, Z_\mathrm{date}^T, E_\mathrm{date}, S_\mathrm{date} \right),
\end{equation}
where $E_\mathrm{date}$ represents the evaluation results for the day, and $S_\mathrm{date}$ is the strategy for that day.

Long-term memory is generated through reflection, as represented by:
\begin{equation}\label{Eq7}
	Z'_\mathrm{date} = \left(S_0, S_1, \dots, S_\mathrm{date} \right).
\end{equation}

The process for updating long-term memory is defined as follows:
\begin{equation}\label{Eq8}
\begin{aligned}
	 & Z'_{\mathrm{date} + 1} = S_\mathrm{date} \cup Z'_\mathrm{date}, \\
	 & \mathrm{date} \in \{0, 1, \dots, \mathrm{DAYS}\}.
\end{aligned} 
\end{equation}
where $Z'_\mathrm{date}$ represents the long-term memory, and $\mathrm{DAYS}$ denotes the maximum number of days.

Together, short-term and long-term memory provides essential context for the agents. Success in this environment depends on their ability to understand the game rules and develop strategies that outmaneuver competitors. Agents continuously refine their strategies based on incremental quantitative feedback, adjusting their actions to align with long-term objectives. 

\section{\textsc{NASDAQ} and \textsc{CSI}}
\label{appendix_NASDAQ}

This study selects seven stocks from the NASDAQ exchange: AAPL, AMZA, GOOGL, MSFT, NFLX, NVDA, and TSLA. These stocks represent leading companies in the technology, energy, and automotive sectors, providing high market representativeness and significant trading volumes. The NASDAQ dataset from Yahoo Finance spans July 3, 2023, to October 29, 2024, excluding weekends and holidays. This dataset reflects current market trends and serves as a timely foundation for our research. It includes daily records of opening price, closing price, highest price, lowest price, and trading volume, as well as relevant technical indicators, offering a comprehensive view of market behavior.

This study also selects seven representative stocks from the Chinese stock market (\textsc{CSI}), covering various sectors with industry-leading companies: Kweichow Moutai (600519.SH), China Merchants Bank (600036.SH), CATL (300750.SZ), BYD (002594.SZ), LONGi Green Energy (601012.SH), ZTE (000063.SZ) and Mindray (300760.SZ). These stocks span consumer goods, finance, new energy, telecommunications, and healthcare, representing leading companies with strong market influence, stable performance, and innovative potential. For \textsc{CSI}, we conducted a one-year simulation from January 2 to December 31, 2024, using the same initial capital setup, while excluding weekends and official trading holidays. The data set is sourced from Baostock and includes daily records such as the opening price, closing price, highest price, lowest price, trading volume, and multiple technical indicators. This comprehensive data set provides a solid foundation for analyzing market trends and investment strategies within the CSI framework.

The training and testing periods for StockFormer~\citep{stockformer}, TimesNet~\citep{timesnet}, StockMixer~\citep{stockmixer}, 
and our system is shown in \cref{real_stock}. StockFormer, TimesNet, and StockMixer require longer training periods, and their training data sets span a broader time range compared to the other models. In contrast, our system does not require training and relies solely on historical stock data to make trading decisions. 

\begin{table*}[h]
	\centering
	\footnotesize
	\setlength{\tabcolsep}{3.5pt}
	\begin{tabular}{l|ccc|cc|ccc|cc}
	\toprule[1.1pt]
	\multirow{3}[4]{*}{Strategy} 
	& \multicolumn{5}{c|}{\textsc{NASDAQ}} 
	& \multicolumn{5}{c}{\textsc{CSI}} \\
	\cmidrule(lr){2-6} \cmidrule(lr){7-11} 
	& \multicolumn{3}{c|}{Training} & \multicolumn{2}{c|}{Testing} & \multicolumn{3}{c|}{Training} & \multicolumn{2}{c}{Testing}\\
	\cmidrule(lr){2-4} \cmidrule(lr){5-6} \cmidrule(lr){7-9} \cmidrule(lr){10-11}
	 & & \multicolumn{1}{c}{Start} & End & Start & End & & \multicolumn{1}{c}{Start} & End & Start & End \\
	\midrule
	Buy \& Hold & \textit{w/o} & 25/8/2024 & 30/8/2024 & 3/9/2024 & 29/10/2024 & \textit{w/o} & 21/12/2023 & 29/12/2023 & 2/1/2024 & 31/12/2024 \\
	SMA & \textit{w/o} & 1/8/2024 & 30/8/2024 & 3/9/2024 & 29/10/2024 & \textit{w/o} & 1/12/2023 & 29/12/2023 & 2/1/2024 & 31/12/2024 \\
	ZMR & \textit{w/o} & 10/8/2024 & 30/8/2024 & 3/9/2024 & 29/10/2024 & \textit{w/o} & 8/12/2023 & 29/12/2023 & 2/1/2024 & 31/12/2024 \\
	MACD & \textit{w/o} & 4/8/2024 & 30/8/2024 & 3/9/2024 & 29/10/2024 & \textit{w/o} & 4/12/2023 & 29/12/2023 & 2/1/2024 & 31/12/2024 \\
	\midrule
	StockFormer & \textit{w/} & 3/7/2023 & 30/8/2024 & 3/9/2024 & 29/10/2024 & \textit{w/} & 2/1/2020 & 29/12/2023 & 2/1/2024 & 31/12/2024 \\
	TimesNet & \textit{w/} & 3/7/2023 & 30/8/2024 & 3/9/2024 & 29/10/2024 & \textit{w/} & 2/1/2020 & 29/12/2023 & 2/1/2024 & 31/12/2024 \\
	StockMixer & \textit{w/} & 3/7/2023 & 30/8/2024 & 3/9/2024 & 29/10/2024 & \textit{w/} & 2/1/2020 & 29/12/2023 & 2/1/2024 & 31/12/2024 \\
	Ours & \textit{w/o} & 16/8/2024 & 30/8/2024 & 3/9/2024 & 29/10/2024 & \textit{w/o} & 1/12/2023 & 29/12/2023 & 2/1/2024 & 31/12/2024 \\
	\bottomrule[1.1pt]
	\end{tabular}
	\caption{\textbf{Training and Testing Periods for Various Models on \textsc{NASDAQ} and \textsc{CSI}.}}
	\label{real_stock}
\end{table*}

The testing period is set from September 3, 2024, to October 29, 2024, to prevent potential data leaks that could provide prior knowledge to the GPT-4o system. This timeframe ensures the fairness of the evaluation by mitigating biases from such leaks, thereby enhancing the reliability of the experimental outcomes.

We also select two stocks from the CSI dataset, each representing a different growth pattern, for simulation. As shown in \cref{real_data_A_share_stock}, our method consistently outperforms the average trend of the selected stocks.

\begin{table}
	\centering
	\footnotesize
	\setlength{\tabcolsep}{5.5pt}
	\begin{tabular}{ccc|lcc}
	\toprule[1.1pt]
	\multicolumn{3}{c|}{Trend} & \multirow{2}[2]{*}{LLMs} & \multirow{2}[2]{*}{TR $\uparrow$} & \multirow{2}[2]{*}{SR $\uparrow$} \\
	\cmidrule(lr){1-3}
	 \multicolumn{1}{c}{Average} & Min & Max \\
	\midrule
	\multirow{3}{*}{-5.86} & \multirow{3}{*}{-13.77} & \multirow{3}{*}{-0.21} & GPT4o & -0.50 & -0.021 \\
	& & & Qwen & \textbf{2.01} & 0.207 \\
	& & & LlaMa3 & 1.44 & 0.122 \\
 
	\midrule
	\multirow{3}{*}{11.37} & \multirow{3}{*}{-4.25} & \multirow{3}{*}{26.05} & GPT4o & \textbf{20.61} & 0.465 \\
	& & & Qwen & 16.46 & 0.420 \\
	& & & LlaMa3 & 4.14 & 0.167 \\
	\bottomrule[1.1pt]
	\end{tabular}
	\caption{\textbf{Performance on CSI Stocks with Different Growth Patterns.} Our method consistently outperforms the average trend of both selected stocks.}
	\label{real_data_A_share_stock}
	%\vspace{-3pt}
\end{table}

\section{Prompt Design Across Modalities}
\label{appendix_prompt} 

The prompts used for stock analysis with text-only, vision-only, and combined inputs are shown below. Additional prompts for others can be found at: \url{https://github.com/wekjsdvnm/Agent-Trading-Arena}.

%\subsection{Text-Only Prompt Design}

% 示例代码片段，插入在正文中
\begin{figure*}[!hpt]
\centering
\begin{tcolorbox}[colback=gray!5, colframe=black!70, width=0.95\linewidth, boxrule=0.5pt, arc=2mm, title=Text-only Prompt Design]
\small
\textbf{Instructions:} 

\noindent\hspace{2em}- You are a player participating in a Simulated Stock Trading Challenge where the prices of stocks are determined mainly by the bids made by the participants in the challenge.

\noindent\hspace{2em}- You will be given information related to specific stocks, the whole market, and your existing investments, which is used to analyze the current stock market and investment situation.

\noindent\hspace{2em}- Your overall objective is to make as much profit as possible. 

\vspace{0.2em}
\textbf{Stock information:}

\noindent\hspace{2em}- A:
        
        \noindent\hspace{4em}- The closing prices in the past 10 days are: [437.80, 459.44, 465.25, 490.38, 501.034, 511.65, 511.72, 511.79, 511.78, 511.79]
        
       \noindent\hspace{4em} - Dividend per share: 22
       
       \noindent\hspace{4em} - Current price change: +0.00\%, Current price: 511.79
       
       \noindent\hspace{4em} - Intraday High: 511.79
       
       \noindent\hspace{4em} - Intraday Low: 511.79
       
       \noindent\hspace{4em}- Intraday Mean: 511.79

    \noindent\hspace{2em}- B:
    
       \noindent\hspace{4em} - The closing prices in the past 10 days are: [460.75, 465.80, 493.27, 502.06, 502.49, 497.32, 486.28, 468.01, 480.61, 480.61]
       
        \noindent\hspace{4em}- Dividend per share: 23
        
        \noindent\hspace{4em} - Current price change: +0.00\%, Current price: 480.61
        
        \noindent\hspace{4em}- Intraday High: 480.61
        
        \noindent\hspace{4em}- Intraday Low: 480.61
        
        \noindent\hspace{4em}- Intraday Mean: 480.61

    \noindent\hspace{2em}- C:
    
        \noindent\hspace{4em}- The closing prices in the past 10 days are: [455.90, 440.532, 424.91, 419.75, 420.48, 421.31, 420.12, 421.09, 435.33, 435.33]
        
        \noindent\hspace{4em}- Dividend per share: 25
        
        \noindent\hspace{4em}- Current price change: +-0.00\%, Current price: 435.33
        
        \noindent\hspace{4em}- Intraday High: 435.33
        
        \noindent\hspace{4em}- Intraday Low: 435.33
        
        \noindent\hspace{4em}- Intraday Mean: 435.33

\textbf{Market information:}

\noindent\hspace{2em}Current market index change: 0.58\%

\textbf{Gossip from other people:}

\noindent\hspace{2em}- Exciting developments in the stock market are stirring, with whispers suggesting a potential merger between companies holding Stock A and B. This merger is rumored to lead to significant synergies, possibly resulting in a stock price surge for both A and B. Consequently, many traders anticipate a rise in demand for these stocks, suggesting a timely investment could yield remarkable returns. Meanwhile, analysts rave about Stock C's steady position and attractive dividends, marking it as a fortress stock that continues to deliver......

\textbf{Existing Investments:}

\noindent\hspace{2em}Your total portfolio balance is 134807.00, you are holding the following stocks:

       \noindent\hspace{2em} - B:
       
            \noindent\hspace{4em}- You have held 162 shares of this stock
            
            \noindent\hspace{4em}- the total portfolio value is 73839.6 the total capital gain is 5.00\% PROFIT
            
            \noindent\hspace{4em}- The prices in the past 10 days are: prices: [460.75, 465.80, 493.27, 502.06, 502.49, 497.32, 486.28, 468.01, 480.61, 480.61]
            
           \noindent\hspace{4em} - current price change: +0.00\%, current price: 480.61, cost price: 455.8

       \noindent\hspace{2em} - A:
       
            \noindent\hspace{4em}- You have held 130 shares of this stock
            
            \noindent\hspace{4em}- the total portfolio value is 60967.4 the total capital gain is 6.00\% PROFIT
            
            \noindent\hspace{4em}- The prices in the past 10 days are: prices: [437.80, 459.44, 465.25, 490.38, 501.034, 511.65, 511.72, 511.79, 511.78, 511.79]
            
            \noindent\hspace{4em}- current price change: +0.00\%, current price: 511.79, cost price: 468.98

\textbf{Investment strategy:}

try to maximize profit.

\textbf{Task:}

\noindent\hspace{2em}- Based on (but not limited to) the following information, you need to summarize and conclude the 3 most important and valuable analysis results from the above information. The analysis results must fully fit the focus of the investment strategy, and each result should describe the relationship between its content and investment strategy. In the next stage, the analysis results will be referred for the decision to buy or sell:

    \noindent\hspace{2em}- Stock information
    
    \noindent\hspace{2em} - Market information
    
    \noindent\hspace{2em}- Your existing investments
    
    \noindent\hspace{2em}- Your investment strategy
    
    \noindent\hspace{2em}- Gossip from other people (Gossip may be real or fake news.)

Output the response to the prompt above in JSON. Each analysis result should be started with "-", and ended with a line break.

Please only provide the response in the following format:

%\begin{quote}
\{"output": "The analysis results: [analysis results]"\}
%\end{quote}
\end{tcolorbox}
%\caption{Illustrative prompt used in the simulated stock trading task.}
%\label{fig:prompt_trading}
\end{figure*}

% 示例代码片段，插入在正文中
\begin{figure*}[!hpt]
\centering
\begin{tcolorbox}[colback=gray!5, colframe=black!70, width=0.95\linewidth, boxrule=0.5pt, arc=2mm, title=Visual-only Prompt Design]
\small
\vspace{0.1em}
\textbf{Instructions:} 

\noindent\hspace{2em}- You are a player participating in a Simulated Stock Trading Challenge where the prices of stocks are determined mainly by the bids made by the participants in the challenge.

\noindent\hspace{2em}- You will be given information related to specific stocks, the whole market, and your existing investments, which is used to analyze the current stock market and investment situation.

\noindent\hspace{2em}- Your overall objective is to make as much profit as possible. 

%\vspace{0.5em}
\textbf{Stock information:}

\noindent
\begin{minipage}[b]{0.32\textwidth}
  \includegraphics[width=\linewidth]{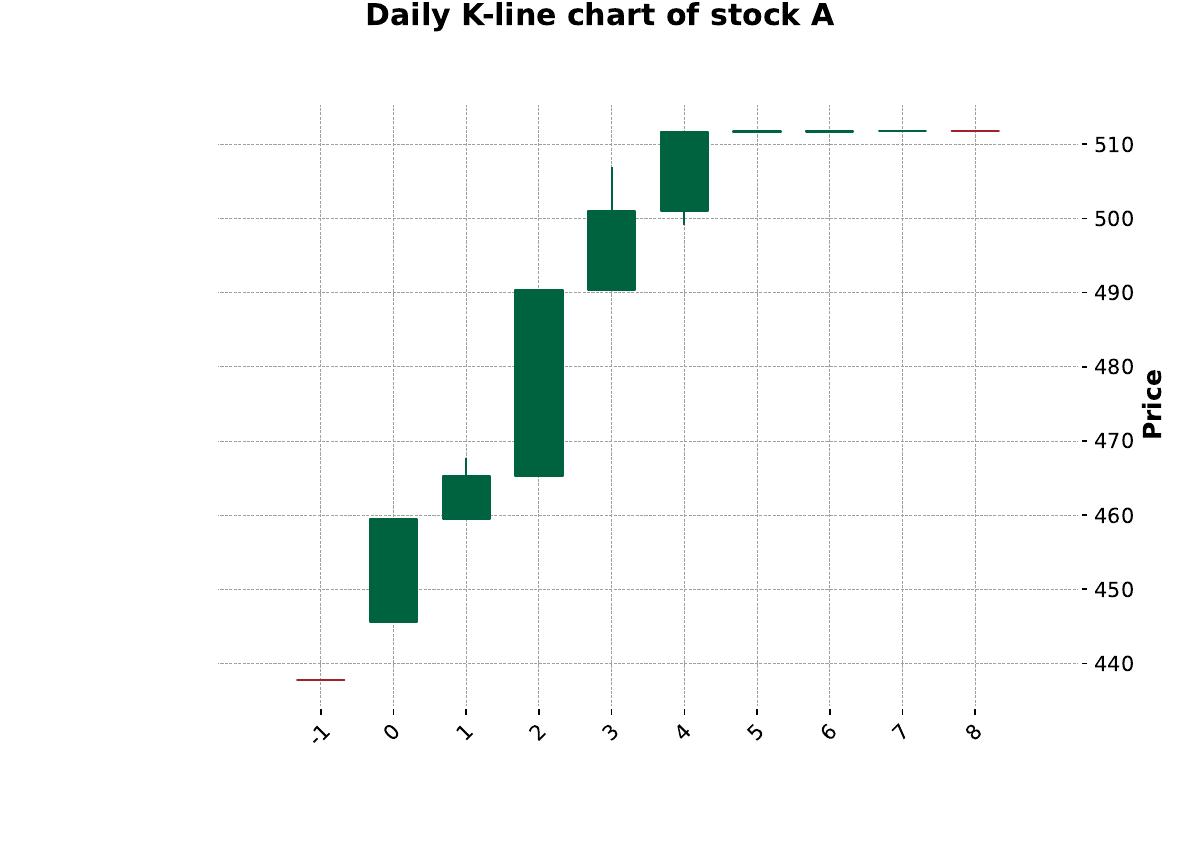}
  \centering (a) Stock A Price
\end{minipage}
\hfill
\begin{minipage}[b]{0.32\textwidth}
  \includegraphics[width=\linewidth]{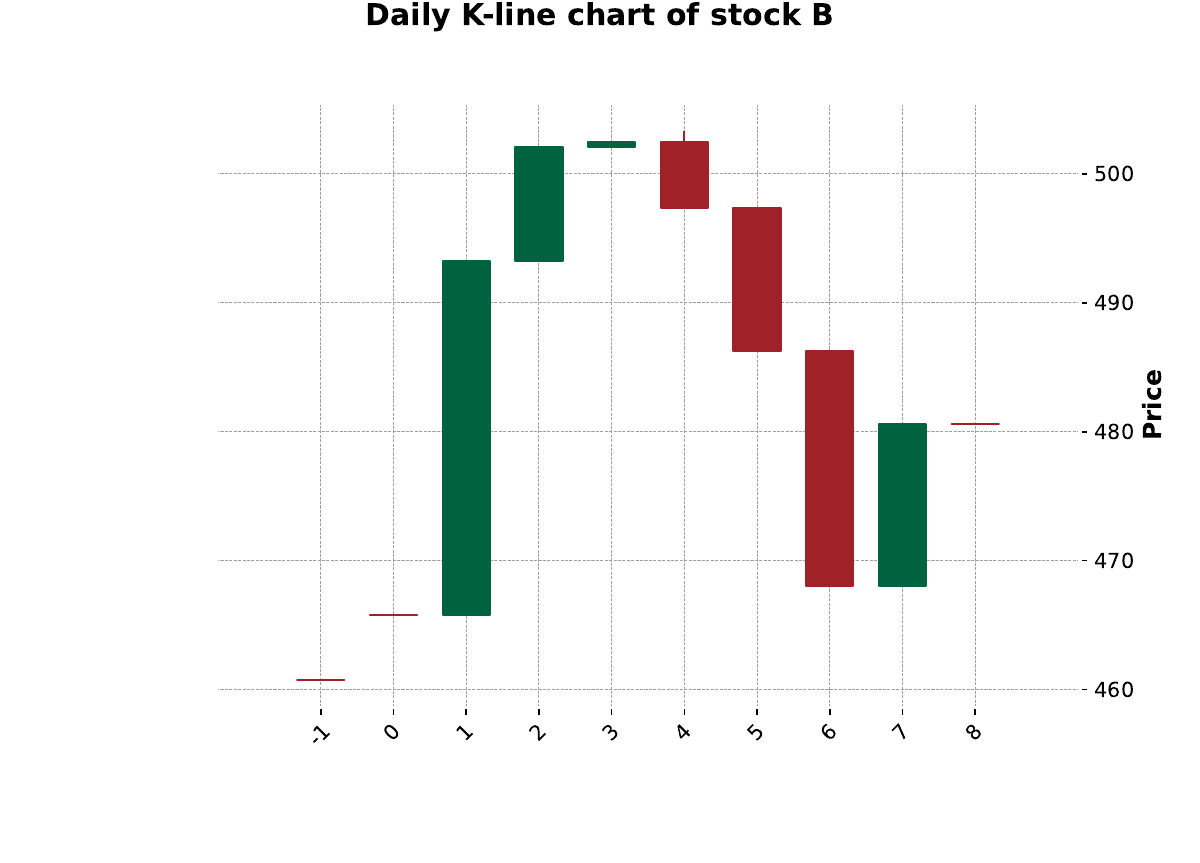}
  \centering (b) Stock B Price
\end{minipage}
\hfill
\begin{minipage}[b]{0.32\textwidth}
  \includegraphics[width=\linewidth]{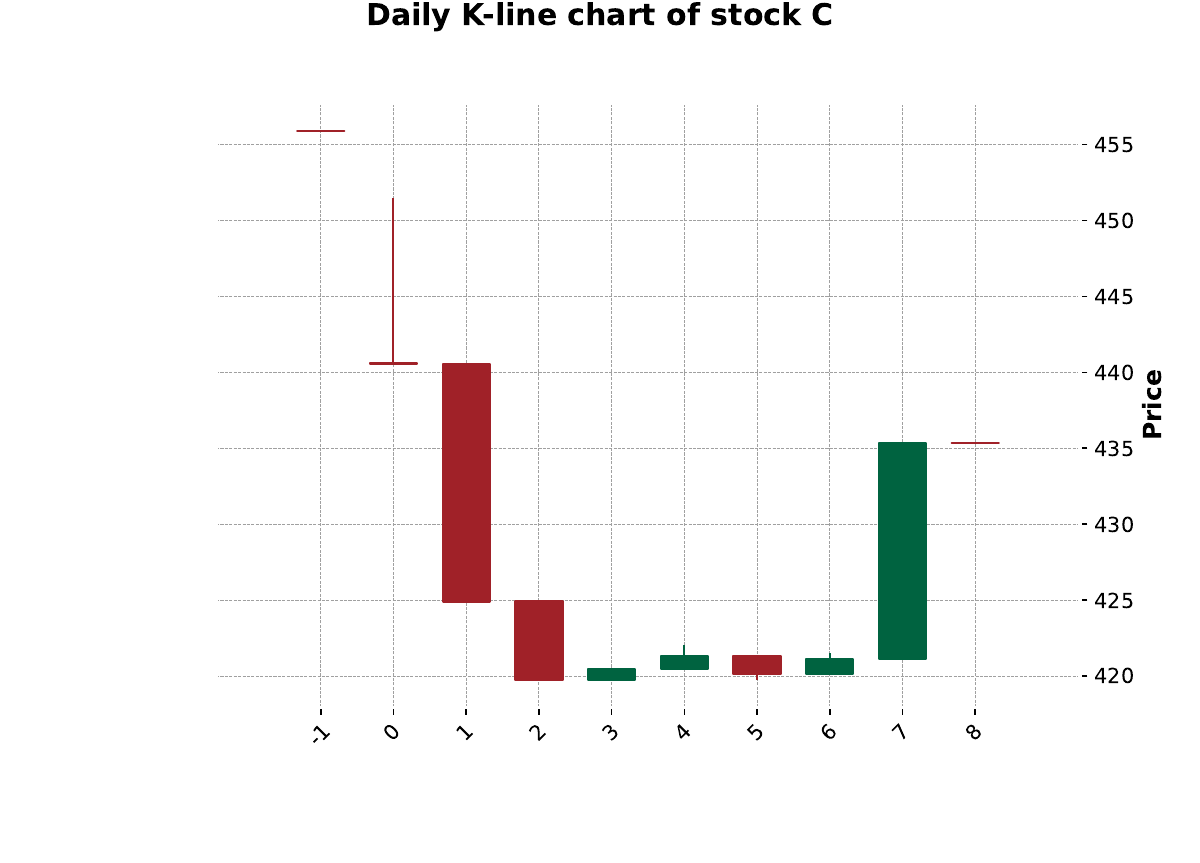}
  \centering (c) Stock C Price
\end{minipage}
\hfill
\begin{minipage}[b]{0.32\textwidth}
  \includegraphics[width=\linewidth]{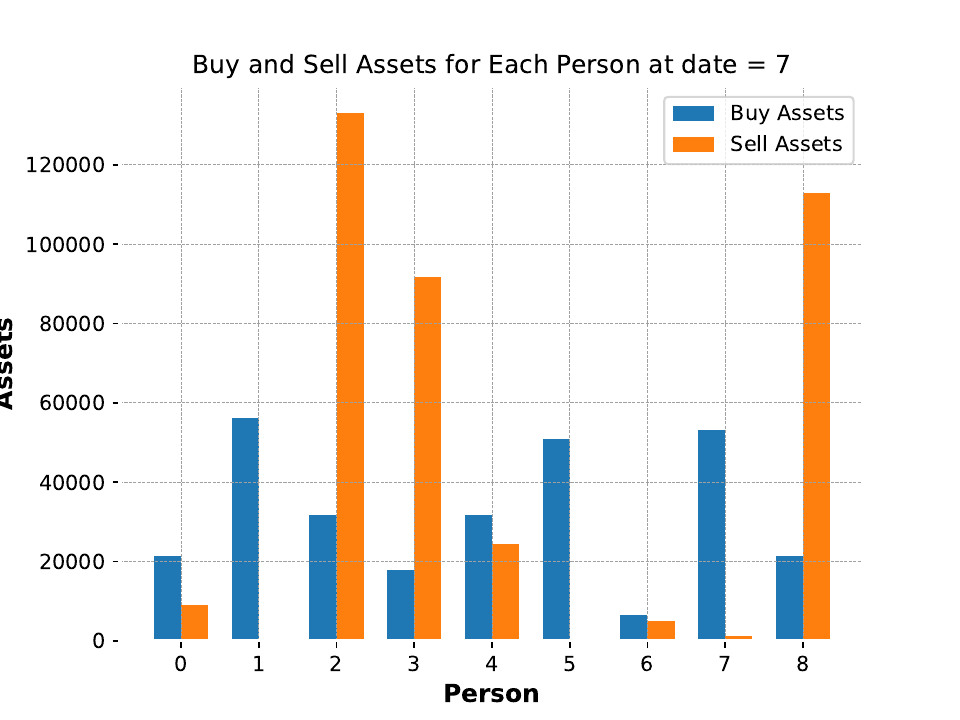}
  \centering (d) Buy \& Sell Assert
\end{minipage}
\hfill
\begin{minipage}[b]{0.32\textwidth}
  \includegraphics[width=\linewidth]{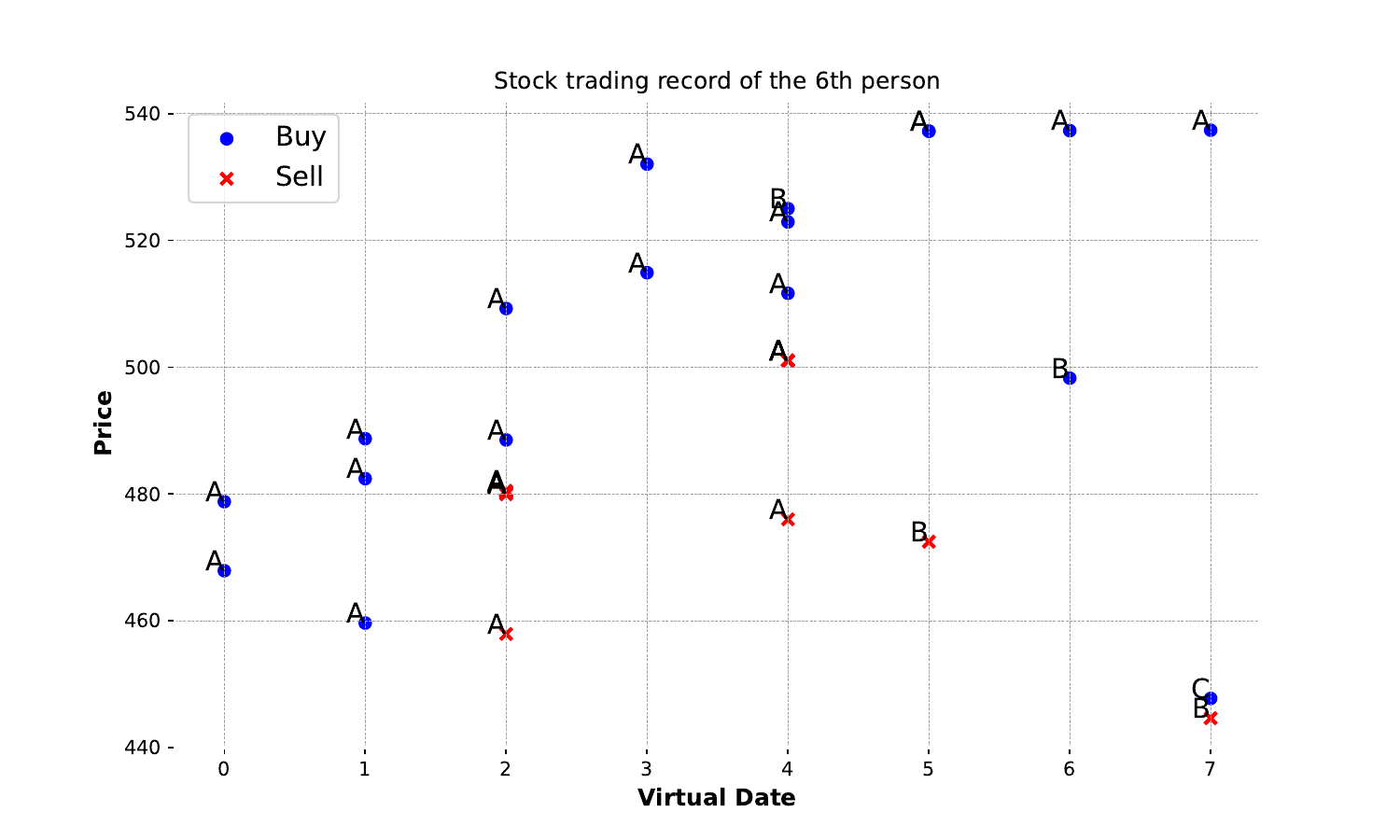}
  \centering (e) Trading Record
\end{minipage}

\textbf{Market information:}

\noindent\hspace{2em}Current market index change: 0.58\%

\textbf{Gossip from other people:}

\noindent\hspace{2em}- Exciting developments in the stock market are stirring, with whispers suggesting a potential merger between companies holding Stock A and B. This merger is rumored to lead to significant synergies, possibly resulting in a stock price surge for both A and B. Consequently, many traders anticipate a rise in demand for these stocks, suggesting a timely investment could yield remarkable returns. Meanwhile, analysts rave about Stock C's steady position and attractive dividends, marking it as a fortress stock that continues to deliver consistent profits with minimal risk. As traders weigh their options, strategic portfolio rebalancing tips towards these dividends-rich stocks seem to be the talk of the town. However, with the volatile nature of stock rumors, it remains essential to tread carefully and keep an eye on reliable sources for confirmation. Will this buzz hold substance, or is it merely hot air? Only time will reveal the truth behind the whispers, leaving the market in suspense.

\textbf{Existing Investments:}

\noindent\hspace{2em}Your total portfolio balance is 134807.00, you are holding the following stocks:

       \noindent\hspace{2em} - C:
       
            \noindent\hspace{4em}- You have held 162 shares of this stock
            
            \noindent\hspace{4em}- the total portfolio value is 73839.6 the total capital gain is 5.00\% PROFIT

       \noindent\hspace{2em} - A:
       
            \noindent\hspace{4em}- You have held 130 shares of this stock
            
            \noindent\hspace{4em}- the total portfolio value is 60967.4 the total capital gain is 6.00\% PROFIT

\textbf{Investment strategy:}

try to maximize profit.

\textbf{Task:}

\noindent\hspace{2em}- Based on (but not limited to) the following information, you need to summarize and conclude the 3 most important and valuable analysis results from the above information. The analysis results must fully fit the focus of the investment strategy, and each result should describe the relationship between its content and investment strategy. In the next stage, the analysis results will be referred for the decision to buy or sell:

    \noindent\hspace{2em}- Stock information
    
    \noindent\hspace{2em} - Market information
    
    \noindent\hspace{2em}- Your existing investments
    
    \noindent\hspace{2em}- Your investment strategy
    
    \noindent\hspace{2em}- Gossip from other people (Gossip may be real or fake news.)

    \noindent\hspace{2em}\textbf{- The provided stock information visualizations.}

Output the response to the prompt above in JSON. Each analysis result should be started with "-", and ended with a line break.

Please only provide the response in the following format:

%\begin{quote}
\{"output": "The analysis results: [analysis results]"\}
%\end{quote}
\end{tcolorbox}
%\caption{Illustrative prompt used in the simulated stock trading task.}
%\label{fig:prompt_trading}
%\vspace{-10pt}
\end{figure*}

% 示例代码片段，插入在正文中
\begin{figure*}[!hpt]
\centering
\begin{tcolorbox}[colback=gray!5, colframe=black!70, width=0.95\linewidth, boxrule=0.5pt, arc=2mm, title=Combined-Input Prompt Design]
\small
\textbf{Instructions:} 

\noindent\hspace{2em}- You are a player participating in a Simulated Stock Trading Challenge where the prices of stocks are determined mainly by the bids made by the participants in the challenge.

\noindent\hspace{2em}- You will be given information related to specific stocks, the whole market, and your existing......% investments, which is used to analyze the current stock market and investment situation.

%\noindent\hspace{2em}- Your overall objective is to make as much profit as possible. 

%\vspace{0.5em}
\textbf{Stock information:}

\noindent\hspace{2em}- A:
        
        \noindent\hspace{4em}- The closing prices in the past 10 days are: [437.80, 459.44, 465.25, 490.38, 501.034, 511.65, 511.72, 511.79, 511.78, 511.79]
        
       \noindent\hspace{4em} - Dividend per share: 22
       
       \noindent\hspace{4em} - Current price change: +0.00\%, Current price: 511.79
       
       \noindent\hspace{4em} - Intraday High: 511.79
       
       \noindent\hspace{4em} - Intraday Low: 511.79
       
       \noindent\hspace{4em}- Intraday Mean: 511.79

    \noindent\hspace{2em}- B:
    
       \noindent\hspace{4em} - The closing prices in the past 10 days are: [460.75, 465.80, 493.27, 502.06, 502.49, 497.32, 486.28, 468.01, 480.61, 480.61]
       
        \noindent\hspace{4em}- Dividend per share: 23
        
        \noindent\hspace{4em} - Current price change: +0.00\%, Current price: 480.61
        
        \noindent\hspace{4em}- Intraday High: 480.61
        
        \noindent\hspace{4em}- Intraday Low: 480.61
        
        \noindent\hspace{4em}- Intraday Mean: 480.61

    \noindent\hspace{2em}- C:
    
        \noindent\hspace{4em}- The closing prices in the past 10 days are: [455.90, 440.532, 424.91, 419.75, 420.48, 421.31, 420.12, 421.09, 435.33, 435.33]
        
        \noindent\hspace{4em}- Dividend per share: 25
        
        \noindent\hspace{4em}- Current price change: +-0.00\%, Current price: 435.33
        
        \noindent\hspace{4em}- Intraday High: 435.33
        
        \noindent\hspace{4em}- Intraday Low: 435.33
        
        \noindent\hspace{4em}- Intraday Mean: 435.33
        
\noindent
\begin{minipage}[b]{0.19\textwidth}
  \includegraphics[width=\linewidth]{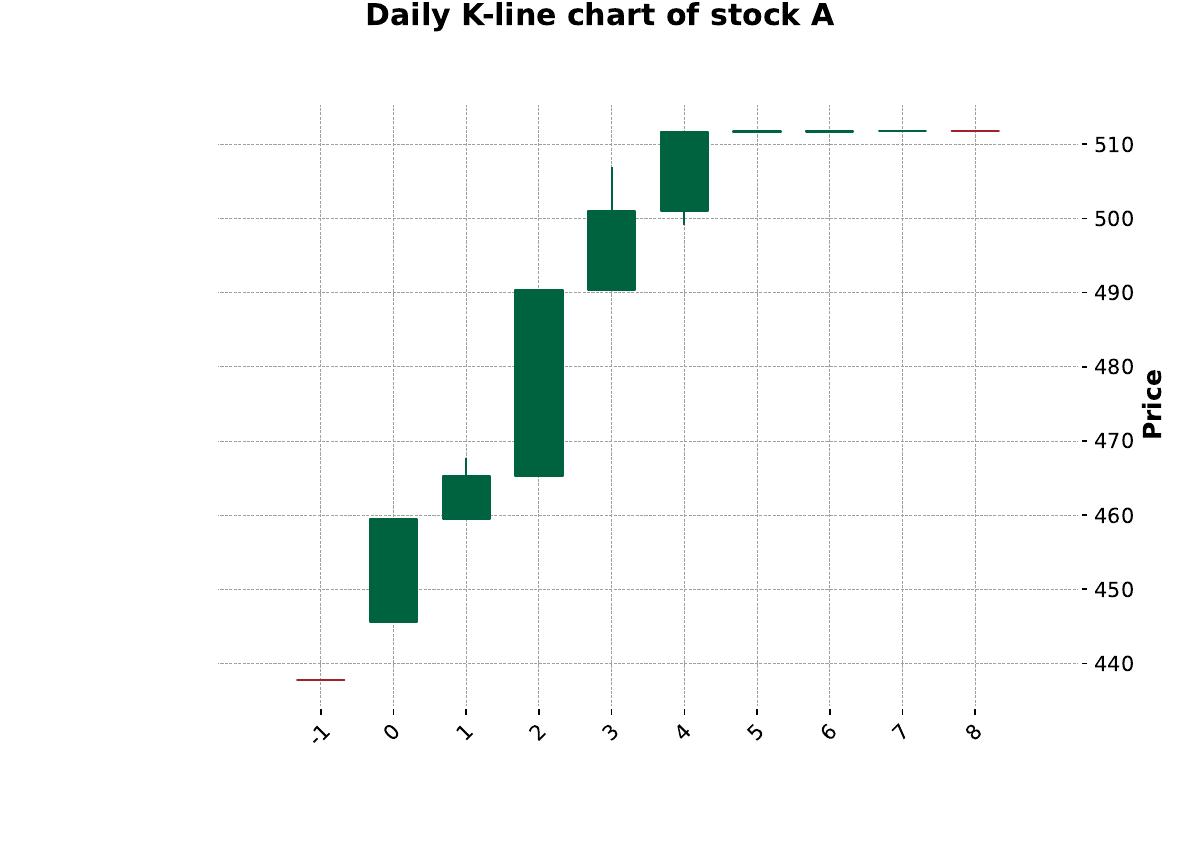}
  \centering (a) Stock A Price
\end{minipage}
\hfill
\begin{minipage}[b]{0.19\textwidth}
  \includegraphics[width=\linewidth]{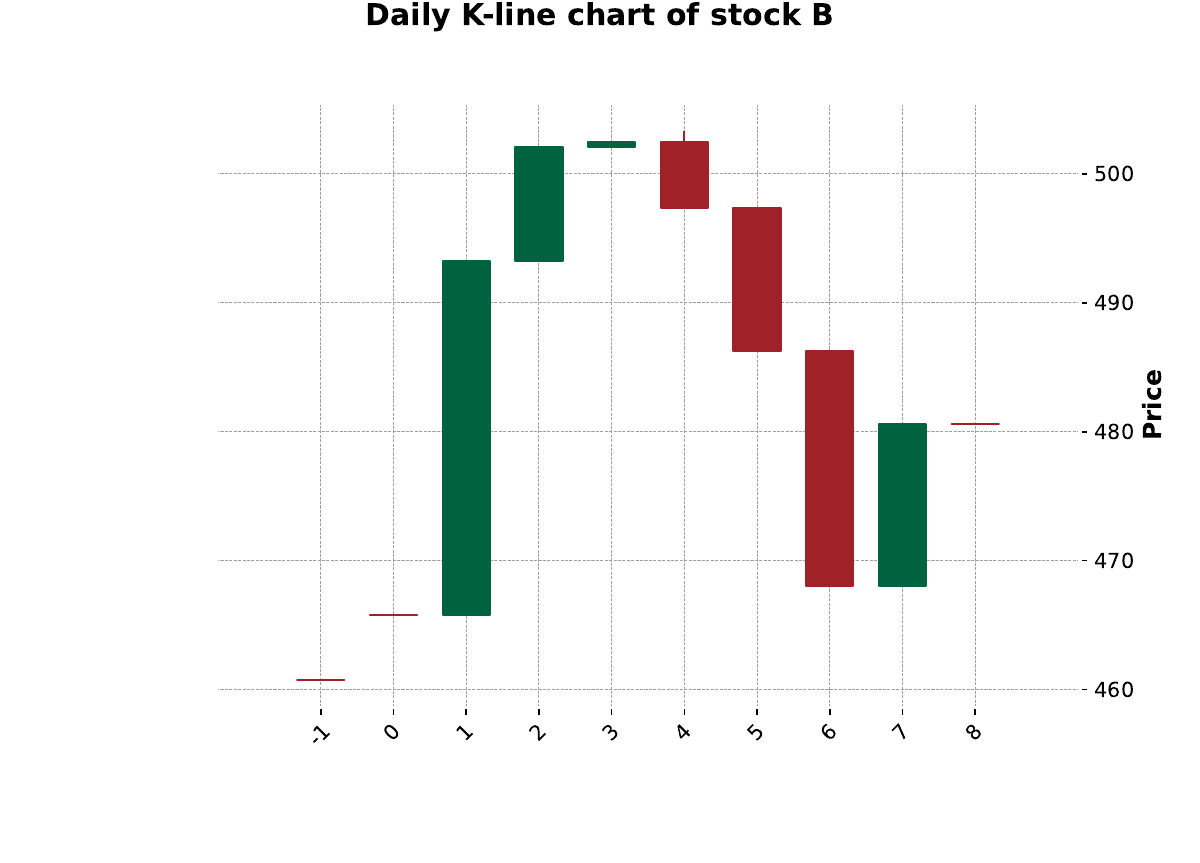}
  \centering (b) Stock B Price
\end{minipage}
\hfill
\begin{minipage}[b]{0.19\textwidth}
  \includegraphics[width=\linewidth]{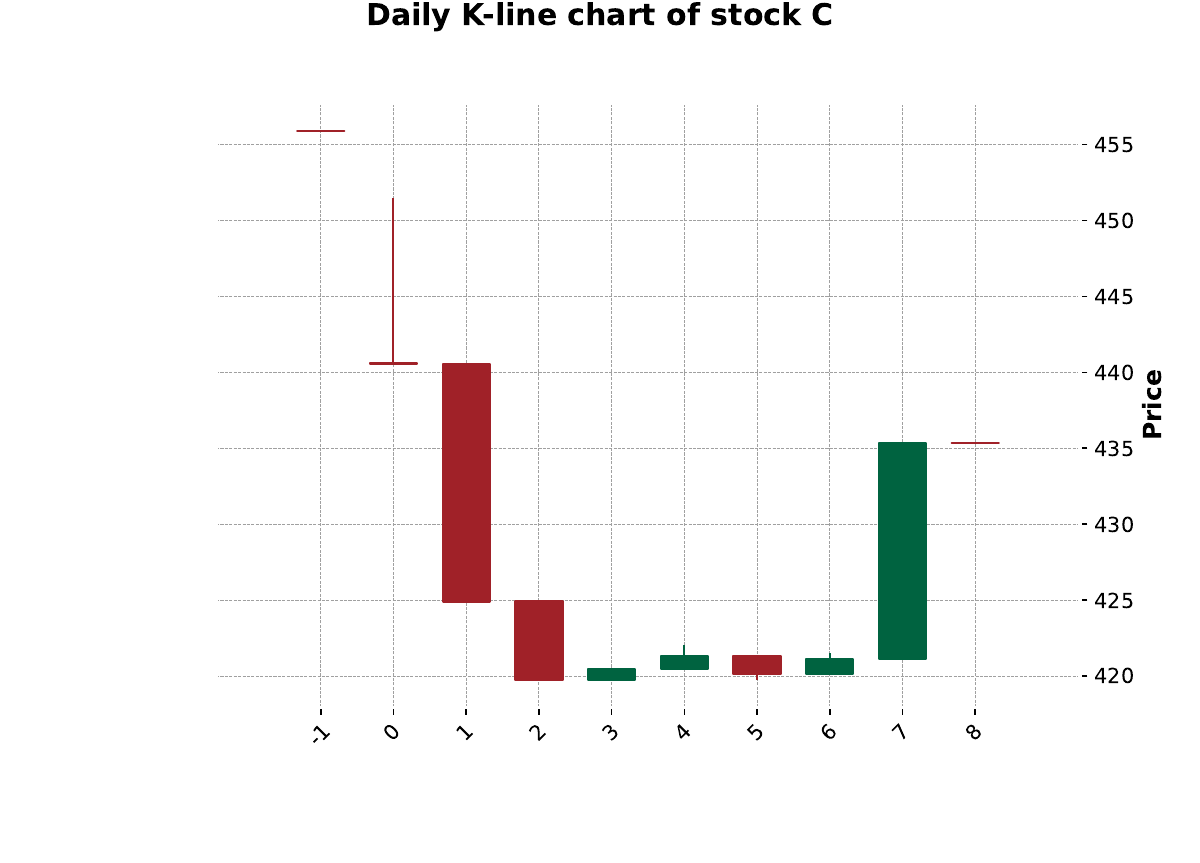}
  \centering (c) Stock C Price
\end{minipage}
\hfill
\begin{minipage}[b]{0.19\textwidth}
  \includegraphics[width=\linewidth]{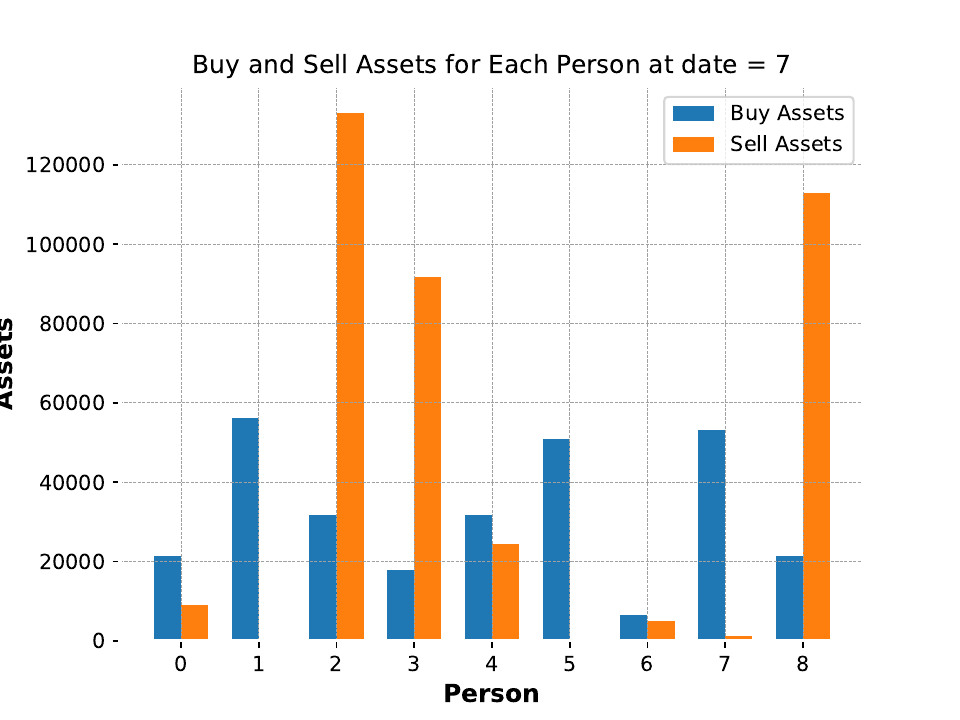}
  \centering (d) Buy \& Sell Assert
\end{minipage}
\hfill
\begin{minipage}[b]{0.19\textwidth}
  \includegraphics[width=\linewidth]{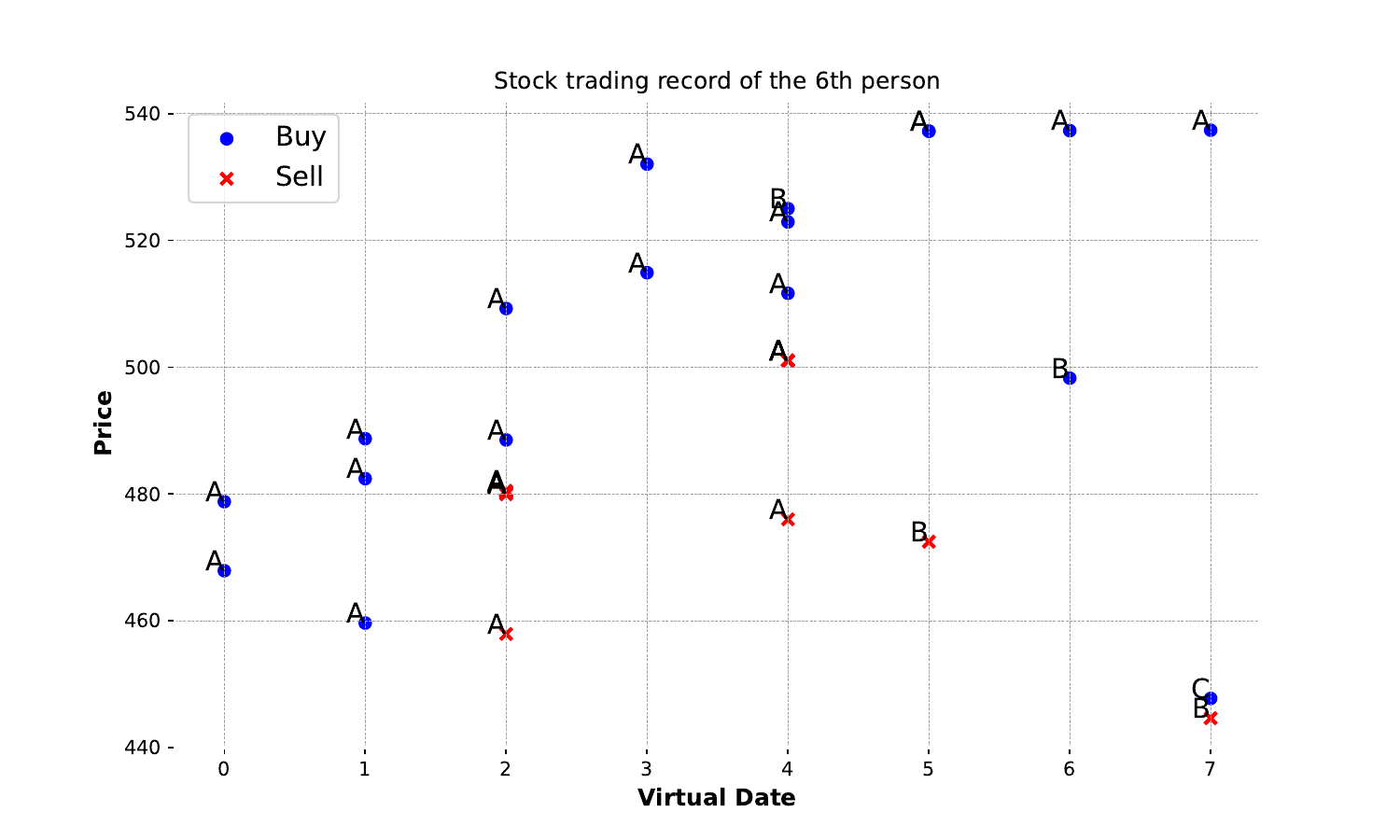}
  \centering (e) Trading Record
\end{minipage}

\textbf{Market information:}

\noindent\hspace{2em}Current market index change: 0.58\%

\textbf{Gossip from other people:}

\noindent\hspace{2em}- Exciting developments in the stock market are stirring, with whispers suggesting a potential merger between companies holding Stock A and B. This merger is rumored to lead to significant synergies......

\textbf{Existing Investments:}

\noindent\hspace{2em}Your total portfolio balance is 134807.00, you are holding the following stocks:

       \noindent\hspace{2em} - B:
       
            \noindent\hspace{4em}- You have held 162 shares of this stock
            
            \noindent\hspace{4em}- the total portfolio value is 73839.6 the total capital gain is 5.00\% PROFIT
            
            \noindent\hspace{4em}- The prices in the past 10 days are: prices: [460.75, 465.80, 493.27, 502.06, 502.49, 497.32, 486.28, 468.01, 480.61, 480.61]
            
           \noindent\hspace{4em} - current price change: +0.00\%, current price: 480.61, cost price: 455.8

       \noindent\hspace{2em} - A:
       
            \noindent\hspace{4em}- You have held 130 shares of this stock
            
            \noindent\hspace{4em}- the total portfolio value is 60967.4 the total capital gain is 6.00\% PROFIT
            
            \noindent\hspace{4em}- The prices in the past 10 days are: prices: [437.80, 459.44, 465.25, 490.38, 501.034, 511.65, 511.72, 511.79, 511.78, 511.79]
            
            \noindent\hspace{4em}- current price change: +0.00\%, current price: 511.79, cost price: 468.98

\textbf{Investment strategy:}

try to maximize profit.

\textbf{Task:}

\noindent\hspace{2em}- Based on (but not limited to) the following information, ...... describe the relationship between its content and investment strategy. In the next stage, the analysis results will be referred for the decision to buy or sell:

    \noindent\hspace{2em}- Stock information
    
    \noindent\hspace{2em} - Market information
    
    \noindent\hspace{2em}- Your existing investments
    
    \noindent\hspace{2em}- Your investment strategy
    
    \noindent\hspace{2em}- Gossip from other people (Gossip may be real or fake news.)

    \noindent\hspace{2em}- The provided stock information visualizations.

Output the response to the prompt above in JSON. Each analysis result should be started with "-",......

Please only provide the response in the following format:

%\begin{quote}
\{"output": "The analysis results: [analysis results]"\}
%\end{quote}
\end{tcolorbox}
%\caption{Illustrative prompt used in the simulated stock trading task.}
%\label{fig:prompt_trading}
\end{figure*}

%\subsection{Visual-Only Prompt Design}

%\subsection{Combined-Input Prompt Design}

%\newpage
\section{Simulation Process}
\label{appendix_LLM}

The experiments involved several LLMs, including LLaMa-3~\citep{LLaMa3}, GPT-4o~\citep{GPT4o}, GPT-4o-mini~\citep{GPT4o}, DeepSeek~\citep{DeepSeek3v}, Qwen-2.5~\citep{qwen2.5}, Qwen-VL-32k~\citep{Qwen2VL}, Qwen-VL-128kk~\citep{Qwen2VL}, and Gemini-1.5~\citep{Gemini1.5}, corresponding to the models Meta-LLaMa-3-70B-Instruct, gpt-4o-2024-08-06, gpt-4o-mini-2024-07-18, DeepSeek-chat, Qwen2.5-72B-Instruct, qwen-vl-max, qwen-vl-max-latest, and Gemini-1.5-pro.

\section{Visualization Input}
\label{appendix_D} 

\begin{figure*}
	\centering
	\includegraphics[width = \linewidth]{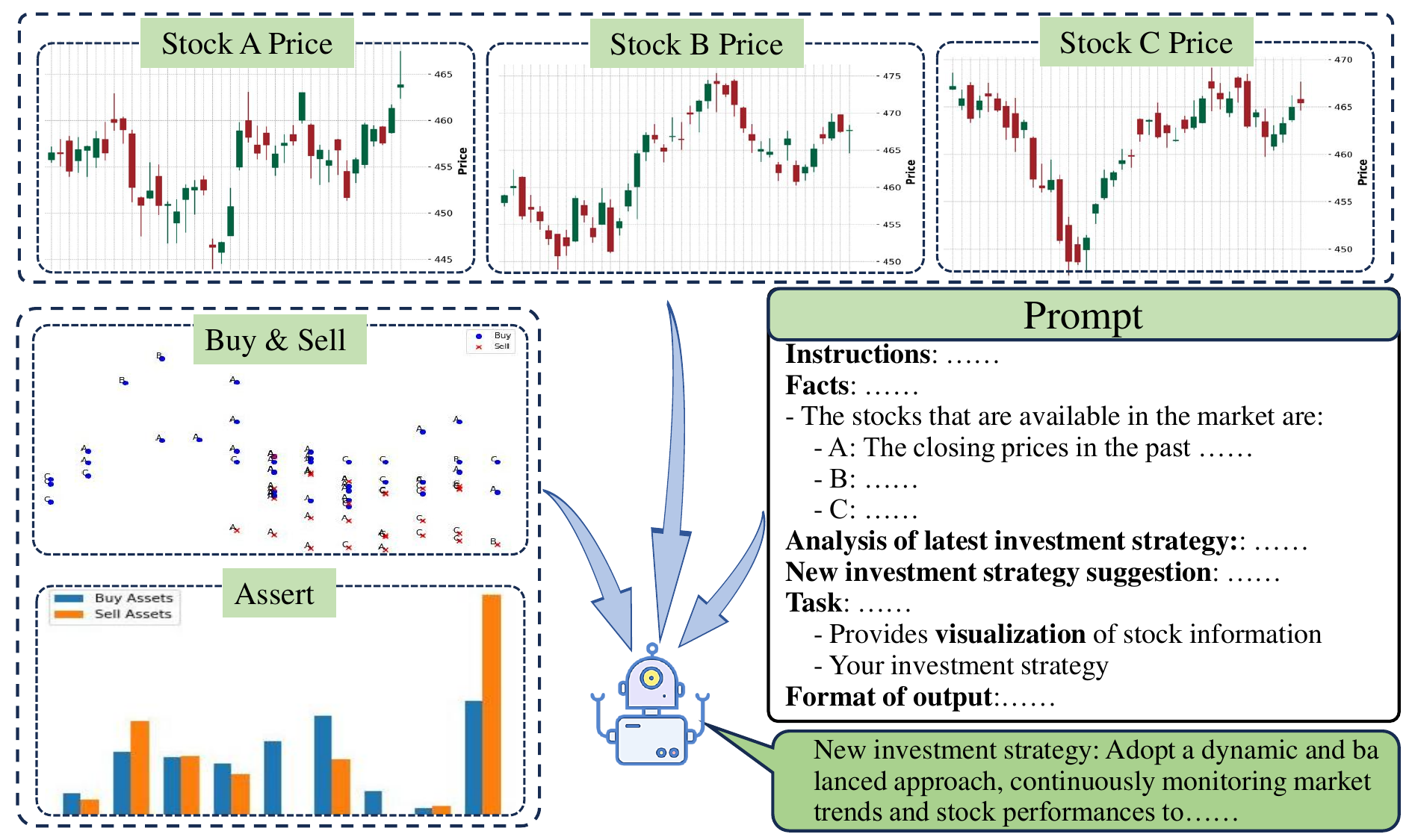}
	\caption{\textbf{Visualization of Stock Inputs and Corresponding Trading Strategy Outputs.}}
	\label{fig:withphoto}
	% % \vspace{-4pt}
\end{figure*}

\cref{fig:withphoto} illustrates the system's input prompts and corresponding outputs during the strategy update process. The input prompts consist of both textual and visual components, including daily K-line charts, transaction histories, and agent trading volumes, all of which inform the strategy update.

\section{Simulation Process}
\label{appendix_E} 

\begin{figure*}
	\centering
	\includegraphics[width = \linewidth]{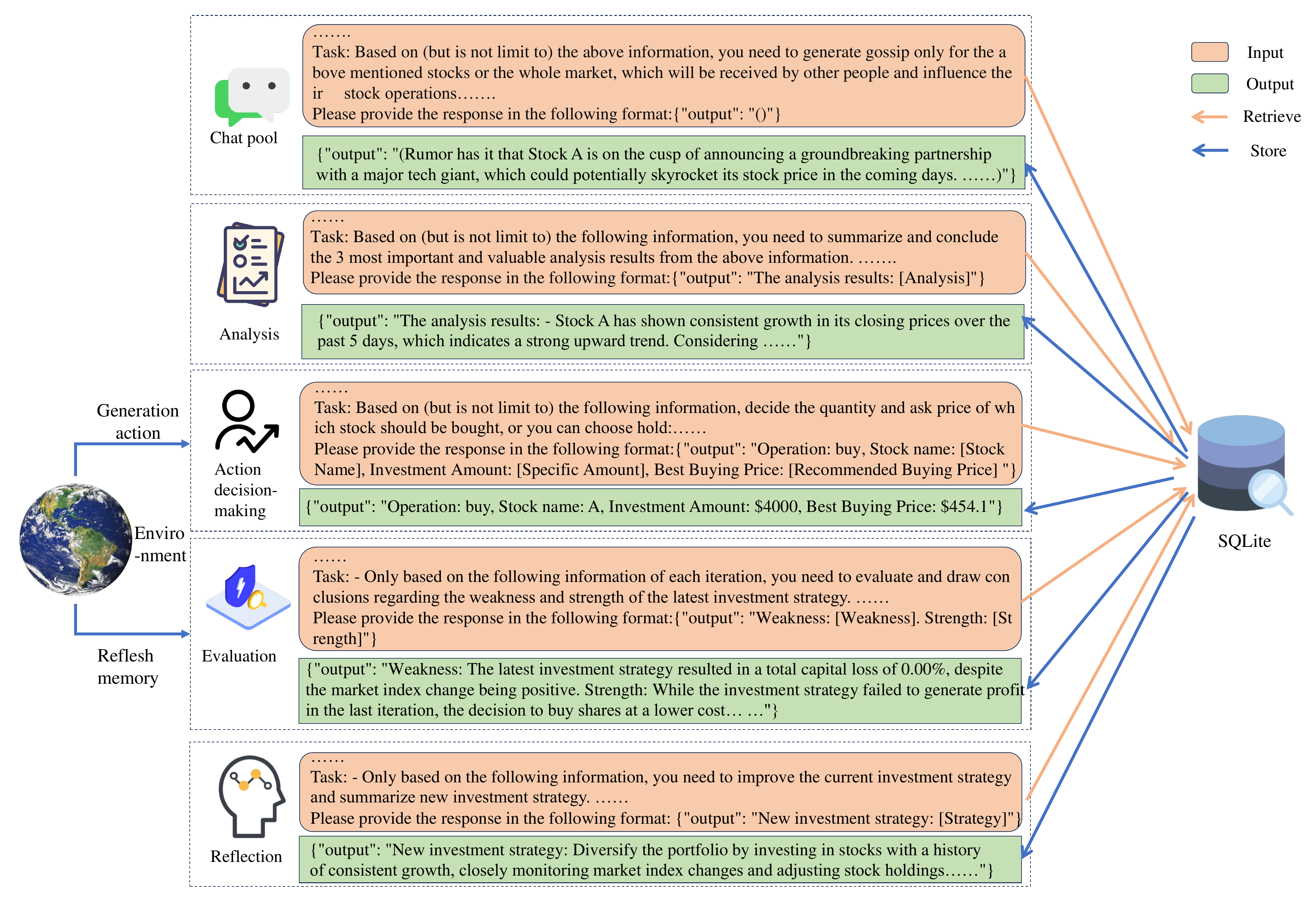}
	\caption{\textbf{Overview of Complete Workflow.}}
	\label{fig: all}
	% % \vspace{-4pt}
\end{figure*}

In the \textit{Agent Trading Arena}, the simulation process unfolds as follows: First, rumors are generated in the chat pool based on the previous day's stock market analysis. Next, historical stock data is analyzed, followed by decision-making and execution. Short-term memory is formed through interactions with the environment. Finally, the system evaluates this memory, updates the strategy, and consolidates it into long-term memory. This entire process is illustrated in \cref{fig: all}.

% \vspace{-4pt}
\paragraph{Impact of Modality on LLM Competitiveness.} 

We employed a relative evaluation method for this experiment. The first and second agents used various LLMs in textual and visual settings, respectively, while the remaining agents were based on LLaMa-3~\citep{LLaMa3} as the baseline. This setup aimed to explore the impact of different agents and modalities on LLM performance. The results are shown in \cref{ablation_experiment_text_vision}. The findings indicate that DeepSeek~\citep{DeepSeek3v} exhibited stronger competitive performance across different LLM environments, suggesting unique strengths or optimizations that enable it to adapt more effectively to the task's complexities.

\begin{table}
	\centering
	\footnotesize
	\setlength{\tabcolsep}{10pt}
	\begin{tabular}{l|cc|cc}
	\toprule[1.1pt]
	LLMs & T & V & \multicolumn{1}{c}{TR $\uparrow$} & \multicolumn{1}{c}{SR $\uparrow$} \\
	\midrule
	LLaMa-3 & \CIRCLE & \Circle & 100.00 & 100.00 \\
	DeepSeek & \CIRCLE & \Circle & \textbf{+ 35.03} & \textbf{+ 45.63} \\
	\midrule
	\multirow{2}{*}{Gemini-1.5} & \CIRCLE & \Circle & + 10.99 & + 14.93 \\
	 & \Circle & \CIRCLE & + 11.49 & + 29.47 \\
	\cmidrule(l){2-5}
	\multirow{2}{*}{GPT-4o} & \CIRCLE & \Circle & + 10.39 & + 28.29 \\
	 & \Circle & \CIRCLE & \underline{+ 17.18} & + \underline{40.77} \\
	\bottomrule[1.1pt]
	\end{tabular}
	\caption{\textbf{Performance Comparison in Trading Decisions \textit{w/o} Reflection, Competing Pairwise with LLaMa-3.} DeepSeek may possess unique strengths or optimizations that allow it to better adapt to the task's complexities when competing with other models.}
	\label{ablation_experiment_text_vision}
	%\vspace{-3pt}
\end{table}

\end{document}